\documentclass{article}

\PassOptionsToPackage{numbers, compress}{natbib}

\usepackage[final]{neurips_2021}

\usepackage[utf8]{inputenc} 
\usepackage[T1]{fontenc}    
\usepackage{hyperref}       
\usepackage{url}            
\usepackage[table]{xcolor}  
\usepackage{amsfonts}       
\usepackage{nicefrac}       
\usepackage{microtype}      
\usepackage{xcolor}         
\usepackage{setspace}

\usepackage{authblk}

\usepackage[rightcaption]{sidecap}
\sidecaptionvpos{figure}{t}
\usepackage{adjustbox}
\usepackage{threeparttable}
\usepackage{etoolbox}
\appto\TPTnoteSettings{\footnotesize}
\usepackage{cellspace}
\usepackage{booktabs}
\usepackage{multirow}
\usepackage{subcaption}
\usepackage{floatrow}
\newfloatcommand{capbtabbox}{table}[][\FBwidth]
\usepackage{algorithm}
\usepackage[noend]{algpseudocode}
\makeatletter
\newcommand{\algmargin}{\the\ALG@thistlm}
\makeatother
\algnewcommand{\parState}[1]{\State%
  \parbox[t]{\dimexpr\linewidth-\algmargin}{\strut {\setstretch{0.5}\hspace{-2.3pt} #1\par
  \vspace{-\prevdepth}
  \vspace{-10pt}
  }\strut}
}
\usepackage{xpatch}
\algrenewcommand\alglinenumber[1]{{\sffamily\scriptsize#1}}
\makeatletter
\xpatchcmd{\algorithmic}{\itemsep\z@}{\itemsep=0pt}{}{}
\makeatother
\usepackage{etoolbox}
\usepackage{tikz}
\usetikzlibrary{tikzmark}
\usetikzlibrary{calc}
\errorcontextlines\maxdimen
\newcommand{\ALGtikzmarkcolor}{black}
\newcommand{\ALGtikzmarkextraindent}{4pt}
\newcommand{\ALGtikzmarkverticaloffsetstart}{-.5ex}
\newcommand{\ALGtikzmarkverticaloffsetend}{-.5ex}
\makeatletter
\newcounter{ALG@tikzmark@tempcnta}
\newcommand\ALG@tikzmark@start{%
    \global\let\ALG@tikzmark@last\ALG@tikzmark@starttext%
    \expandafter\edef\csname ALG@tikzmark@\theALG@nested\endcsname{\theALG@tikzmark@tempcnta}%
    \tikzmark{ALG@tikzmark@start@\csname ALG@tikzmark@\theALG@nested\endcsname}%
    \addtocounter{ALG@tikzmark@tempcnta}{1}%
}
\def\ALG@tikzmark@starttext{start}
\newcommand\ALG@tikzmark@end{%
    \ifx\ALG@tikzmark@last\ALG@tikzmark@starttext
    \else
        \tikzmark{ALG@tikzmark@end@\csname ALG@tikzmark@\theALG@nested\endcsname}%
        \tikz[overlay,remember picture] \draw[\ALGtikzmarkcolor] let \p{S}=($(pic cs:ALG@tikzmark@start@\csname ALG@tikzmark@\theALG@nested\endcsname)+(\ALGtikzmarkextraindent,\ALGtikzmarkverticaloffsetstart)$), \p{E}=($(pic cs:ALG@tikzmark@end@\csname ALG@tikzmark@\theALG@nested\endcsname)+(\ALGtikzmarkextraindent,\ALGtikzmarkverticaloffsetend)$) in (\x{S},\y{S})--(\x{S},\y{E});%
    \fi
    \gdef\ALG@tikzmark@last{end}%
}
\apptocmd{\ALG@beginblock}{\ALG@tikzmark@start}{}{\errmessage{failed to patch}}
\pretocmd{\ALG@endblock}{\ALG@tikzmark@end}{}{\errmessage{failed to patch}}
\makeatother
\usepackage{amsmath}
\usepackage{amssymb}
\usepackage{amsthm}
\usepackage{amsfonts}
\usepackage{mathtools}
\usepackage{bm}
\usepackage{bbm}
\usepackage{stmaryrd}
\usepackage{mathrsfs}

\usepackage{amsmath,amsfonts,bm}

\newcommand{\x}{\bm{x}} 

\def\eqref#1{Eqn.~\ref{#1}}

\def\eqref#1{equation~\ref{#1}}

\def\1{\bm{1}}

\def\vone{{\bm{1}}}

\def\va{{\bm{a}}}

\def\vc{{\bm{c}}}

\def\ve{{\bm{e}}}

\DeclareMathAlphabet{\mathsfit}{\encodingdefault}{\sfdefault}{m}{sl}
\SetMathAlphabet{\mathsfit}{bold}{\encodingdefault}{\sfdefault}{bx}{n}

\def\gL{{\mathcal{L}}}

\def\gN{{\mathcal{N}}}

\def\gR{{\mathcal{R}}}

\def\sR{{\mathbb{R}}}

\def\sZ{{\mathbb{Z}}}

\usepackage{xargs}
\usepackage[capitalize]{cleveref}
\usepackage{paralist}
\usepackage[shortlabels]{enumitem}
\newtheoremstyle{break}
  {}
  {}
  {\itshape}
  {}
  {\bfseries}
  {}
  {\newline}
  {}

\newtheorem{theorem}{Theorem}
\newtheorem{lemma}{Lemma}
\newtheorem{corollary}[theorem]{Corollary}
\newtheorem{innercustomgeneric}{\customgenericname}
\providecommand{\customgenericname}{}
\newcommand{\newcustomtheorem}[2]{%
  \newenvironment{#1}[1]
  {%
   \renewcommand\customgenericname{#2}%
   \renewcommand\theinnercustomgeneric{##1}%
   \innercustomgeneric
  }
  {\endinnercustomgeneric}
}
\newcustomtheorem{customtheorem}{Theorem}
\newcustomtheorem{customlemma}{Lemma}
\newcustomtheorem{customcorollary}{Corollary}
\newenvironment{prevproof}[1]{\noindent {\em {Proof of \cref{#1}:}}}{\hfill $\square$\vskip \belowdisplayskip}
\newcommand{\cm}{\paragraph}

\usepackage[disable]{todonotes}
\newcounter{todocounter}
\newcommandx{\plan}[2][1=]{\stepcounter{todocounter}\todo[linecolor=blue,backgroundcolor=blue!25,bordercolor=blue,#1]{\thetodocounter: #2}}
\newcommandx{\change}[2][1=]{\stepcounter{todocounter}\todo[linecolor=red,backgroundcolor=red!25,bordercolor=red,#1]{\thetodocounter: #2}}
\newcommandx{\unsure}[2][1=]{\stepcounter{todocounter}\todo[linecolor=yellow,backgroundcolor=yellow!25,bordercolor=yellow,#1]{\thetodocounter: #2}}
\newcommandx{\info}[2][1=]{\stepcounter{todocounter}\todo[linecolor=green,backgroundcolor=green!25,bordercolor=green,#1]{\thetodocounter: #2}}
\makeatletter
\renewcommand*\env@matrix[1][\arraystretch]{%
  \edef\arraystretch{#1}%
  \hskip -\arraycolsep
  \let\@ifnextchar\new@ifnextchar
  \array{*\c@MaxMatrixCols c}}
\makeatother
\makeatletter
\newcommand*\bigcdot{\mathpalette\bigcdot@{.5}}
\newcommand*\bigcdot@[2]{\mathbin{\vcenter{\hbox{\scalebox{#2}{$\m@th#1\bullet$}}}}}
\makeatother
\renewcommand{\emph}[1]{\textit{#1}}
\newcommand{\func}[1]{\textsc{\textbf{#1}}}
\newcommand{\wt}[1]{\widetilde{#1}}
\newcommand{\grad}[1]{\nabla_{#1}\ell}
\newcommand{\apx}[1]{\widehat{#1}}
\newcommand{\apxgrad}[1]{\widehat{\nabla}_{#1}\ell}
\newcommand{\diag}[1]{\mathrm{diag}(#1)}
\newcommand{\diaginv}[1]{\mathrm{diag}^{-1}(#1)}

\newcommand{\idk}[1]{\mathbbm{1}\{#1\}}
\newcommand{\concat}{\bigparallel}

\newcommand{\tran}{^{\mkern-1.5mu\mathsf{T}}}
\newcommand{\transs}{\mkern-1.5mu\mathsf{T}}
\newcommand{\seq}[1]{\left\langle #1\right\rangle}
\newcommand{\exps}[1]{\mathrm{e}^{#1}}
\newcommand{\rank}[1]{\mathrm{rank}\left(#1\right)}
\newcommand{\nonlinear}{\sigma}
\newcommand{\within}{_{\mathrm{in}}}
\newcommand{\outof}{_{\mathrm{out}}}
\newcommand{\fC}{\mathfrak{C}}
\newcommand{\aC}{\mathscr{C}}
\newcommand{\ba}{\bm{\eta}}
\newcommand{\bb}{\bm{\Sigma}}
\newcommand{\itand}{~~{\footnotesize\textsf{and}}~~}
\newcommand{\lip}{\gL ip}
\newcommand{\siunit}[2]{$#1$\texttt{ #2}}

\title{
    VQ-GNN: A Universal Framework to Scale-up\\
    Graph Neural Networks using Vector Quantization
}

\author[ ]{Mucong Ding\thanks{Equal contribution.} }
\author[ ]{Kezhi Kong\protect\footnotemark[1] }
\author[ ]{Jingling Li}
\author[ ]{Chen Zhu}
\author[ ]{\\John Dickerson}
\author[ ]{Furong Huang}
\author[ ]{Tom Goldstein}
\affil[ ]{Department of Computer Science, University of Maryland}
\affil[ ]{\textit {\{mcding, kong, jingling, chenzhu, john, furongh, tomg\}@cs.umd.edu}}

\begin{document}

\maketitle


\begin{abstract}
  Most state-of-the-art Graph Neural Networks (GNNs) can be defined as a form of graph convolution which can be realized by message passing between direct neighbors or beyond. To scale such GNNs to large graphs, various neighbor-, layer-, or subgraph-sampling techniques are proposed to alleviate the ``neighbor explosion'' problem by considering only a small subset of messages passed to the nodes in a mini-batch. However, sampling-based methods are difficult to apply to GNNs that utilize many-hops-away or global context each layer, show unstable performance for different tasks and datasets, and do not speed up model inference. We propose a principled and fundamentally different approach, VQ-GNN, a universal framework to scale up any convolution-based GNNs using Vector Quantization (VQ) without compromising the performance. In contrast to sampling-based techniques, our approach can effectively preserve all the messages passed to a mini-batch of nodes by learning and updating a small number of quantized reference vectors of global node representations, using VQ within each GNN layer. Our framework avoids the ``neighbor explosion'' problem of GNNs using quantized representations combined with a low-rank version of the graph convolution matrix. We show that such a compact low-rank version of the gigantic convolution matrix is sufficient both theoretically and experimentally. In company with VQ, we design a novel approximated message passing algorithm and a nontrivial back-propagation rule for our framework. Experiments on various types of GNN backbones demonstrate the scalability and competitive performance of our framework on large-graph node classification and link prediction benchmarks.
\end{abstract}


\section{Introduction}
\label{sec:intro}


The rise of Graph Neural Networks (GNNs) has brought the modeling of complex graph data into a new era. Using message-passing, GNNs iteratively share information between neighbors in a graph to make predictions of node labels, edge labels, or graph-level properties. A number of powerful GNN architectures~\citep{kipf2016semi,hamilton2017inductive,velivckovic2017graph,xu2018powerful} have been widely applied to solve down-stream tasks such as recommendation, social analysis, visual recognition, etc.

With the soaring size of realistic graph datasets and the industrial need to model them efficiently, GNNs are hindered by a scalability problem. An $L$-layer GNN aggregates information from all $L$-hop neighbors, and standard training routines require these neighbors to all lie on the GPU at once.  This prohibits full-batch training when facing a graph with millions of nodes~\citep{hu2020open}.


\begin{figure}[t]
    \centering
    \includegraphics[width=1.0\textwidth]{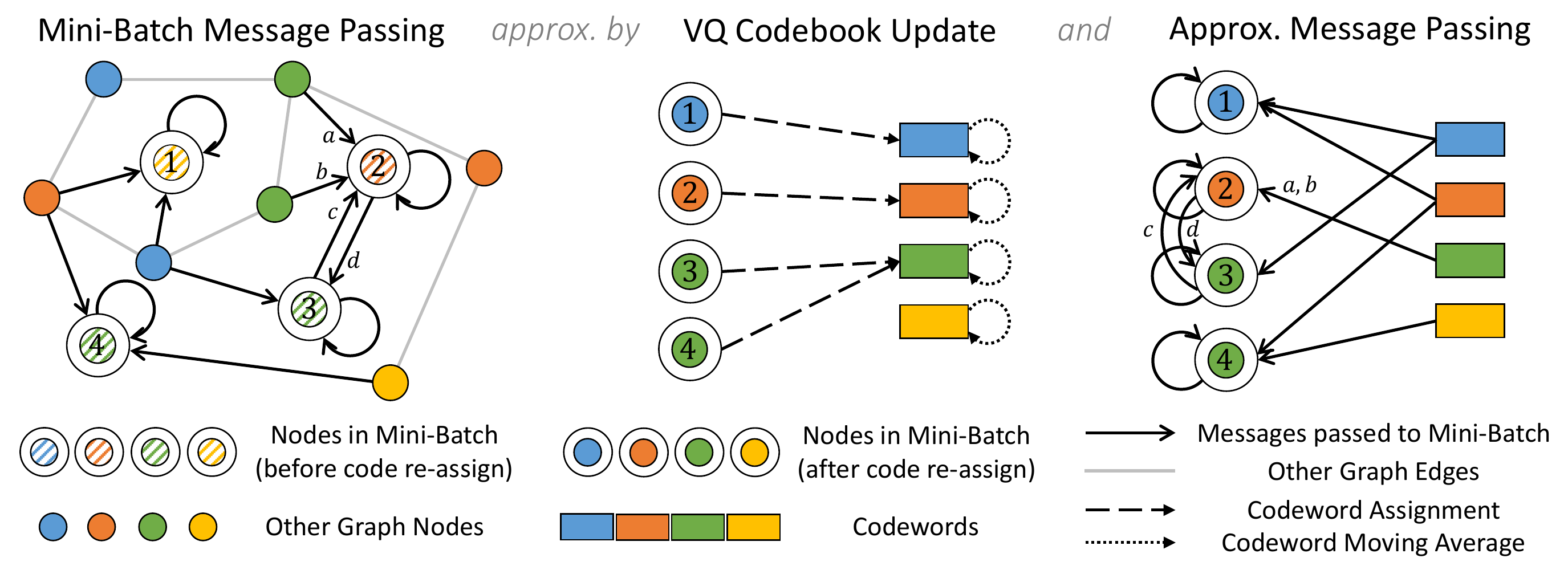}
    \caption{In our framework, VQ-GNN, each \emph{mini-batch message passing} (left) is approximated by a \emph{VQ codebook update} (middle) and an \emph{approximated message passing} (right). All the messages passed to the nodes in the current mini-batch are effectively preserved. Circles are nodes, and rectangles are VQ codewords. A double circle indicates nodes in the current mini-batch. Color represents codeword assignment. During \emph{VQ codebook update}, codeword assignment of nodes in the mini-batch is refreshed (node $1$), and codewords are updated using the assigned nodes. During \emph{approximated message passing}, messages from out-of-mini-batch nodes are approximated by messages from the corresponding codewords, messages from nodes assigned to the same codeword are merged ($a$ and $b$), and intra-mini-batch messages are not changed ($c$ and $d$).}
    \label{fig:schema-graph}
    \vspace{-5pt}
\end{figure}


A number of sampling-based methods have been proposed to accommodate large graphs with limited GPU resources. These techniques can be broadly classified into three categories: (1) Neighbor-sampling methods~\citep{hamilton2017inductive,chen2018stochastic} sample a fixed-number of neighbors for each node; (2) Layer-sampling methods~\citep{chen2018fastgcn,zou2019layer} sample nodes in each layer independently with a constant sample size; (3) Subgraph-sampling methods~\citep{chiang2019cluster,zeng2019graphsaint} sample a subgraph for each mini-batch and perform forward and back-propagation on the same subgraph across all layers. Although these sampling-based methods may significantly speed up the training time of GNNs, they suffer from the following three major drawbacks: (1) At inference phase, sampling methods require all the neighbors to draw non-stochastic predictions, resulting in expensive predictions if the full graph cannot be fit on the inference device; (2) As reported in~\citep{hu2020open} and in~\cref{sec:experiments}, state-of-the-art sampling-baselines fail to achieve satisfactory results consistently across various tasks and datasets; (3) Sampling-based methods cannot be universally applied to GNNs that utilize many-hop or global context in each layer, which hinders the application of more powerful GNNs to large graphs.


This paper presents VQ-GNN, a GNN framework using vector quantization to scale most state-of-the-art GNNs to large graphs through a principled and fundamentally different approach compared with the sampling-based methods. We explore the idea of using vector quantization (VQ) as a means of dimensionality reduction to learn and update a small number of quantized reference vectors (codewords) of global node representations. In VQ-GNN, mini-batch message passing in each GNN layer is approximated by a VQ codebook update and an approximated form of message passing between the mini-batch of nodes and codewords; see~\cref{fig:schema-graph}. Our approach avoids the ``neighbor explosion'' problem and enables mini-batch training and inference of GNNs. In contrast to sampling-based techniques, VQ-GNN can effectively preserve all the messages passed to a mini-batch of nodes. We theoretically and experimentally show that our approach is efficient in terms of memory usage, training/inference time, and convergence speed. Experiments on various GNN backbones demonstrate the competitive performance of our framework compared with the full-graph training baseline and sampling-based scalable algorithms.


\cm{Paper organization.}
The remainder of this paper is organized as follows. \cref{sec:background} summarizes GNNs that can be re-formulated into a common framework of graph convolution. \cref{sec:insight} defines the scalability challenge of GNNs and shows that dimensionality reduction is a potential solution. In \cref{sec:method}, we describe our approach, VQ-GNN, from theoretical framework to algorithm design and explain why it solves the scalability issue of most GNNs. \cref{sec:related} compares our approach to the sampling-based methods. \cref{sec:experiments} presents a series of experiments that validate the efficiency, robustness, and universality of VQ-GNN. Finally, \cref{sec:conclusion} concludes this paper with a summary of limitations and broader impacts.

\section{Preliminaries: GNNs defined as Graph Convolution}
\label{sec:background}


\cm{Notations.}
Consider a graph with $n$ nodes and $m$ edges (average degree $d=m/n$). Connectivity is given by the adjacency matrix $A\in\{0,1\}^{n\times n}$ and features are defined on nodes by $X\in\sR^{n\times f_0}$ with $f_0$ the length of feature vectors. Given a matrix $C$, let $C_{i,j}$, $C_{i,:}$, and  $C_{:,j}$ denote its $(i,j)$-th entry, $i$-th row, $j$-th column, respectively. For a finite sequence $\seq{i_b}: i_1, \ldots, i_b$, we use $C_{\seq{i_b}, :}$ to denote the matrix whose rows are the $i_b$-th rows of matrix $C$. We use $\odot$ to denote the element-wise (Hadamard) product. $\|\cdot\|_p$ denotes the entry-wise $\ell^p$ norm of a vector and $\|\cdot\|_F$ denotes the Frobenius norm. We use $I_n\in\sR^{n\times n}$ to denote the identity matrix, $\vone_n\in\sR^n$ to denote the vector whose entries are all ones, and $\ve^i_n$ to denote the unit vector in $\sR^{n}$ whose $i$-th entry is $1$. The $0$-$1$ indicator function is $\idk{\cdot}$. We use $\diag{\vc}$ to denote a diagonal matrix whose diagonal entries are from vector $\vc$. And $\concat$ represents concatenation along the last axis. We use superscripts to refer to different copies of same kind of variable. For example, $X^{(l)}\in\sR^{n\times f_l}$ denotes node representations on layer $l$. A Graph Neural Network (GNN) layer takes the node representation of a previous layer $X^{(l)}$ as input and produces a new representation $X^{(l+1)}$, where $X=X^{(0)}$ is the input features.


\begin{table}[t]
\centering
\caption{\label{tab:GNN-models}Summary of GNNs re-formulated as generalized graph convolution.}
\adjustbox{max width=1.0\textwidth}{%
\begin{threeparttable}
\renewcommand*{\arraystretch}{1.4}
\begin{tabular}{Sc Sc Sc Sc Sl} \toprule
Model Name & Design Idea & Conv. Matrix Type & \# of Conv. & Convolution Matrix \\ \midrule
GCN\tnote{1}~~\citep{kipf2016semi} & Spatial Conv. & Fixed & $1$ & $C=\wt{D}^{-1/2}\wt{A}\wt{D}^{-1/2}$ \\
SAGE-Mean\tnote{2}~~\citep{hamilton2017inductive} & Message Passing & Fixed & $2$ & \multicolumn{1}{l}{$\left\{\begin{tabular}[c]{@{}l@{}} $C^{(1)}=I_n$\\ $C^{(2)}=D^{-1}A$ \end{tabular}\right.\kern-\nulldelimiterspace$} \\
GAT\tnote{3}~~\citep{velivckovic2017graph} & Self-Attention & Learnable & \# of heads & \multicolumn{1}{l}{$\left\{\begin{tabular}[c]{@{}l@{}} $\fC^{(s)}=A+I_n$\itand\\ $h^{(s)}_{\va^{(l,s)}}(X^{(l)}_{i,:}, X^{(l)}_{j,:})=\exp\big(\mathrm{LeakyReLU}($\\ \quad$(X^{(l)}_{i,:}W^{(l,s)}\concat X^{(l)}_{j,:}W^{(l,s)})\cdot\va^{(l,s)})\big)$ \end{tabular}\right.\kern-\nulldelimiterspace$} \\ \bottomrule
\end{tabular}
\begin{tablenotes}[para]
    \item[1] Where $\wt{A}=A+I_n$, $\wt{D}=D+I_n$.
    \item[2] $C^{(2)}$ represents mean aggregator. Weight matrix in~\citep{hamilton2017inductive} is $W^{(l)}=W^{(l,1)}\concat W^{(l,2)}$.
    \item[3] Need row-wise normalization. $C^{(l,s)}_{i,j}$ is non-zero if and only if $A_{i,j}=1$, thus GAT follows direct-neighbor aggregation.
\end{tablenotes}
\end{threeparttable}}
\vspace{-5pt}
\end{table}


\textbf{A common framework for generalized graph convolution.}
Finally, use a table to summarize popular types of GNN models into the general graph convolution framework and refer readers to the appendix for more details.
Although many GNNs are designed following different guiding principles including\change{I have more citations, but need to decide whether to put here.} neighborhood aggregation (GraphSAGE~\citep{hamilton2017inductive}, PNA~\citep{corso2020principal})\unsure{Double check whether PNA can fit into the framework.}, spatial convolution (GCN~\citep{kipf2016semi}), spectral filtering (ChebNet~\citep{defferrard2016convolutional}, CayleyNet~\citep{levie2018cayleynets}, ARMA~\citep{bianchi2021graph}\unsure{Double check whether ARMA can fit into the framework.}), self-attention (GAT~\citep{velivckovic2017graph}, Graph Transformers~\citep{yaronlipman2020global, rong2020self, zhang2020graph}\unsure{Or should say transformers on graphs?}\unsure{Remove last citation here?}), diffusion (GDC~\citep{klicpera2019diffusion}, DCNN~\citep{atwood2016diffusion})\unsure{The more well-known name of GDC is ``diffusion improves graph learning''.}, Weisfeiler-Lehman (WL) alignment (GIN~\citep{xu2018powerful}, 3WL-GNNs~\citep{morris2019weisfeiler, maron2019provably})\unsure{Be careful about the names of 3WL-GNNs.}, or other graph algorithms (\citep{xu2019can, loukas2019graph})\unsure{Check the last citation later.}.  Despite these differences, {\em nearly all GNNs can be interpreted as performing message passing on node features, followed by feature transformation and an activation function}. As pointed out by~\citet{balcilar2021analyzing}, GNNs can typically be written in the form
\begin{equation}
\label{eq:gnn-forward}
    X^{(l+1)} = \nonlinear\left(\sum_{s} C^{(s)}X^{(l)}W^{(l,s)}\right),
\end{equation}
where $C^{(s)}\in\sR^{n\times n}$ denotes the $s$-th convolution matrix that defines the message passing operator, $s\in\sZ_+$ denotes index of convolution, and $\nonlinear(\cdot)$ denotes the non-linearity. $W^{(l,s)}\in\sR^{f_l\times f_{l+1}}$ is the learnable linear weight matrix for the $l$-th layer and $s$-th filter. 

Within this common framework, GNNs differ from each other by choice of convolution matrices $C^{(s)}$, which can be either fixed or learnable. A learnable convolution matrix relies on the inputs and learnable parameters and can be different in each layer (thus denoted as $C^{(l,s)}$):
\begin{equation}
\label{eq:learnable-conv}
    C^{(l,s)}_{i,j}= \underbrace{\fC^{(s)}_{i,j}}_{\text{fixed}} \cdot \underbrace{h^{(s)}_{\theta^{(l,s)}}(X^{(l)}_{i,:}, X^{(l)}_{j,:})}_{\text{learnable}}
\end{equation}
where $\fC^{(s)}$ denotes the fixed mask of the $s$-th learnable convolution, which may depend on the adjacency matrix $A$ and input edge features $E_{i,j}$. While $h^{(s)}(\cdot, \cdot):\sR^{f_l}\times\sR^{f_l}\to\sR$ can be any learnable model parametrized by $\theta^{(l,s)}$. Sometimes a learnable convolution matrix may be further row-wise normalized as $C^{(l,s)}_{i,j} \gets C^{(l,s)}_{i,j}/\sum_j C^{(l,s)}_{i,j}$, for example in GAT~\citep{velivckovic2017graph}. We stick to~\cref{eq:learnable-conv} in the main paper and discuss row-wise normalization in~\cref{apd:generalized-conv,apd:algorithm}. The receptive field of a layer of graph convolution (\cref{eq:gnn-forward}) is defined as a set of nodes $\gR^1_i$ whose features $\{X^{(l)}_{j,:}\mid j\in\gR_i\}$ determines $X^{(l+1)}_{i,:}$. We re-formulate some popular GNNs into this generalized graph convolution framework; see~\cref{tab:GNN-models} and~\cref{apd:generalized-conv} for more.

The back-propagation rule of GNNs defined by~\cref{eq:gnn-forward} is as follows,
\begin{equation}
\label{eq:gnn-backward}
    \grad{X^{(l)}} = \sum_{s}\left(C^{(l,s)}\right)\tran\bigg(\grad{X^{(l+1)}}\odot\nonlinear'\Big(\nonlinear^{-1}\big(X^{(l+1)}\big)\Big)\bigg)\left(W^{(l,s)}\right)\tran,
\end{equation}
which can also be understood as a form of message passing. $\nonlinear'$ and $\nonlinear^{-1}$ are the derivative and inverse of $\nonlinear$ respectively and $\grad{X^{(l+1)}}\odot\nonlinear'\big(\nonlinear^{-1}(X^{(l+1)})\big)$ is the gradients back-propagated through the non-linearity.

\section{Scalability Problem and Theoretical Framework}
\label{sec:insight}


When a graph is large, we are forced to mini-batch the graph by sampling a subset of $b\ll n$ nodes in each iteration. Say the node indices are $i_1, \ldots, i_b$ and a mini-batch of node features is denoted by $X_B = X_{\seq{i_b},:}$. To mini-batch efficiently for any model, we hope to fetch $\Theta(b)$ information to the training device, spend $\Theta(Lb)$ training time per iteration while taking $(n/b)$ iterations to traverse through the entire dataset. However, it is intrinsically difficult for most of the GNNs to meet these three scalability requirements at the same time. The receptive field of $L$ layers of graph convolution (\cref{eq:gnn-forward}) is recursively given by $\gR^L_i=\bigcup_{j\in\gR^1_i}\gR^{L-1}_j$ (starting with $\gR^1_i\supseteq \{i\}\cup\gN_i$), and its size grows exponentially with $L$. Thus, to optimize on a mini-batch of $b$ nodes, we require $\Omega(bd^L)$ inputs and training time per iteration. Sampling a subset of neighbors~\citep{hamilton2017inductive, chen2018stochastic} for each node in each layer does not change the exponential dependence on $L$. Although layer-~\citep{chen2018fastgcn, huang2018adaptive} and subgraph-sampling~\citep{chiang2019cluster, zeng2019graphsaint} may require only $\Omega(b)$ inputs and $\Omega(Lb)$ training time per iteration, they are only able to consider an exponentially small proportion of messages compared with full-graph training. Most importantly, all existing sampling methods do not support dense convolution matrices with $O(n^2)$ non-zero terms. Please see~\cref{sec:related} for a detailed comparison with sampling-based scalable methods after we introduce our framework.


\cm{Idea of dimensionality reduction.}
We aim to develop a scalable algorithm for any GNN models that can be re-formulated as~\cref{eq:gnn-forward}, where the convolution matrix can be either fixed or learnable, and either sparse or dense. The major obstacle to scalability is that, for each layer of graph convolution, to compute a mini-batch of forward-passed features $X^{(l+1)}_B=X^{(l+1)}_{\seq{i_b},:}$, we need $O(n)$ entries of $C^{(l,s)}_B=C^{(l,s)}_{\seq{i_b},:}$ and $X^{(l)}$, which will not fit in device memory. 

\emph{Our goal is to apply a dimensionality reduction to both convolution and node feature matrices, and then apply convolution using compressed ``sketches'' of $C^{(l,s)}_B$ and $X^{(l)}.$} More specifically, we look for a projection matrix $R\in\sR^{n\times k}$ with $k\ll n$, such that the product of low-dimensional sketches $\wt{C}^{(l,s)}_B=C^{(l,s)}_B R\in\sR^{b\times k}$ and $\wt{X}^{(l)}=R\tran X^{(l)}\in\sR^{k\times f_l}$ is approximately the same as $C^{(l,s)}_BX^{(l)}$. The approximated product (of all nodes) $\wt{C}^{(l,s)}\wt{X}^{(l)}=C^{(l,s)}RR\tran X^{(l)}$ can also be regarded as the result of using a low-rank approximation $C^{(l,s)}RR\tran\in\sR^{n\times n}$ of the convolution matrix such that $\rank{C^{(l,s)}RR\tran}\leq k$. The distributional Johnson–Lindenstrauss lemma~\citep{johnson1984extensions} (JL for short) shows the existence of such projection $R$ with $m=\Theta(\log(n))$, and the following result by~\citet{kane2014sparser} shows that $R$ can be chosen to quite sparse:
\begin{theorem}
\label{thm:jl-lemma}
    For any convolution matrix $C\in\sR^{n\times n}$, any column vector $X_{:,a}\in\sR^{n}$ of the node feature matrix $X\in\sR^{n\times f}$ (where $a=1,\ldots, f$) and $\varepsilon>0$, there exists a projection matrix $R\in\sR^{n\times k}$ (drawn from a distribution) with only an $O(\varepsilon)$-fraction of entries non-zero, such that
    \begin{equation}
    \label{eq:jl-lemma}
        \Pr\left(\|CRR\tran X_{:,a}-CX_{:,a}\|_2<\varepsilon\|CX_{:,a}\|_2\right)>1-\delta,
    \end{equation}
    with $k=\Theta(\log(n)/\varepsilon^2)$ and $\delta=O(1/n)$.
\end{theorem}
Now, the sketches $\wt{C}^{(l,s)}_B$ and $\wt{X}^{(l)}$ take up $O(b\log(n))$ and $\Theta(f_l\log(n))$ memory respectively and can fit into the training and inference device. The sparsity of projection matrix $R$ is favorable because:(1) if the convolution matrix $C^{(l,s)}$ is sparse (e.g., direct-neighbor message passing where only $O(d/n)$-fraction of entries are non-zero), only an $O(\varepsilon d)$-fraction of entries are non-zero in the sketch $\wt{C}^{(l,s)}$; (2) During training, $\wt{X}^{(l)}$ is updated in a ``streaming'' fashion using each mini-batch's inputs $X^{(l)}_B$, and a sparse $R$ reduces the computation time by a factor of $O(\varepsilon)$. However, the projection $R$ produced following the sparse JL-lemma~\citep{kane2014sparser} is randomized and requires $O(\log^2(n))$ uniform random bits to sample. It is difficult to combine this with the deterministic feed-forward and back-propagation rules of neural networks, and there is no clue when and how we should update the projection matrix. Moreover, randomized projections destroy the ``identity'' of each node, and for learnable convolution matrices (\cref{eq:learnable-conv}), it is impossible to compute the convolution matrix only using the sketch of features $\wt{X}^{(l)}$. For this idea to be useful, we need a deterministic and identity-preserving construction of the projection matrix $R\in\sR^{n\times k}$ to avoid these added complexities.

\section{Proposed Method: Vector Quantized GNN}
\label{sec:method}


\cm{Dimensionality reduction using Vector Quantization (VQ).}
A natural and widely-used method to reduce the dimensionality of data in a deterministic and identity-preserving manner is Vector Quantization~\citep{gray1998quantization} (VQ), a classical data compression algorithm that can be formulated as the following optimization problem:
\begin{equation}
\label{eq:vector-quantization}
    \min_{R\in\{0,1\}^{n\times k}\text{, }\wt{X}\in\sR^{k\times f}}\|X-R\wt{X}\|_F \quad \text{s.t.  } R_{i,:}\in\{\ve^1_k,\ldots,\ve^k_k\},
\end{equation}
which is classically solved via \emph{k-means}~\citep{gray1998quantization}. Here the sketch of features $\wt{X}$ is called the feature ``codewords.'' $R$ is called the codeword assignment matrix, whose rows are unit vectors in $\sR^{k}$, i.e., $R_{i,v}=1$ if and only if the $i$-th node is assigned to the $v$-th cluster in \emph{k-means}. The objective in~\cref{eq:vector-quantization} is called Within-Cluster Sum of Squares (WCSS), and we can define the relative error of VQ as $\epsilon=\|X-R\wt{X}\|_F/\|X\|_F$. The rows of $\wt{X}$ are the $k$ codewords (i.e., centroids in \emph{k-means}), and can be computed as $\wt{X}=\diaginv{R\tran\vone_n}R\tran X$, which is slightly different from the definition in~\cref{sec:insight} as a row-wise normalization of $R\tran$ is required. The sketch of the convolution matrix $\wt{C}$ can still be computed as $\wt{C}=CR$. In general, VQ provides us a principled framework to learn the low-dimensional sketches $\wt{X}$ and $\wt{C}$, in a deterministic and node-identity-preserving manner. However, to enable mini-batch training and inference of GNNs using VQ, three more questions need to be answered:
\begin{itemize}[leftmargin=*, topsep=1.5pt]
\setlength\itemsep{0.75pt}
    \item How to approximate the forward-passed mini-batch features of nodes using the learned codewords?
    \item How to back-propagate through VQ and estimate the mini-batch gradients of nodes?
    \item How to update the codewords and assignment matrix along with the training of GNN?
\end{itemize}
In the following part of this section, we introduce the VQ-GNN algorithm by answering all the three questions and presenting a scalability analysis.


\cm{Approximated forward and backward message passing.}
To approximate the forward pass through a GNN layer (\cref{eq:gnn-forward}) with a mini-batch of nodes $\seq{i_b}$, we can divide the messages into two categories: intra-mini-batch messages, and messages from out-of-mini-batch nodes; see the right figure of~\cref{fig:schema-graph}. Intra-mini-batch messages $C^{(l,s)}\within X^{(l)}_B$ can always be computed exactly, where $C^{(l,s)}\within=(C^{(l,s)}_B)_{:,\seq{i_b}}\in\sR^{b\times b}$, because they only rely on the previous layer's node features of the current mini-batch. Equipped with the codewords $\wt{X}^{(l)}$ and the codeword assignment of all nodes $R^{(l)}$, we can approximate the messages from out-of-mini-batch nodes as $\wt{C}^{(l,s)}\outof\wt{X}^{(l)}$, where $\wt{X}^{(l)}=\diaginv{R\tran\vone_n}R\tran X^{(l)}$ as defined above and $\wt{C}^{(l,s)}\outof=C^{(l,s)}\outof R$. Here, $C^{(l,s)}\outof$ is the remaining part of the convolution matrix after removing the intra-mini-batch messages, thus $(C^{(l,s)}\outof)_{:,j}=(C^{(l,s)}_B)_{:,j}\idk{j\in\seq{i_b}}$ for any $j\in\{1,\ldots,n\}$, and $\wt{C}^{(l,s)}$ is the sketch of $C^{(l,s)}\outof$. In general, we can easily approximate the forward-passed mini-batch features $X^{(l+1)}_B$ by $\apx{X}^{(l+1)}_B = \nonlinear\big(\sum_{s}(C^{(l,s)}\within X^{(l)}_B + \wt{C}^{(l,s)}\outof\wt{X}^{(l)})W^{(l,s)}\big)$.

However, the above construction of $\apx{X}^{(l+1)}_B$ does not allow us to back-propagate through VQ straightforwardly using chain rules. During back-propagation, we aim at approximating the previous layer's mini-batch gradients $\grad{X^{(l)}_B}$ given the gradients of the (approximated) output $\grad{\apx{X}^{(l+1)}_B}$ (\cref{eq:gnn-backward}). Firstly, we do not know how to compute the partial derivative of $\wt{C}^{(l,s)}\outof$ and $\wt{X}^{(l)}$ with respect to $X^{(l)}_B$, because the learning and updating of VQ codewords and assignment are \emph{data dependent} and are usually realized by an iterative optimization algorithm. Thus, we need to go through an iterative computation graph to evaluate the partial derivative of $R^{(l)}$ with respect to $X^{(l)}_B$, which requires access to many historical features and gradients, thus violating the scalability constraints. Secondly, even if we apply some very rough approximation during back-propagation as in~\citep{oord2017neural}, that is, assuming that the partial derivative of $R^{(l)}$ with respect to $X^{(l)}_B$ can be ignored (i.e., the codeword assignment matrix is detached from the computation graph, known as ``straight through'' back-propagation), we are not able to evaluate the derivatives of codewords $\wt{X}^{(l)}$ because they rely on some node features out of the current mini-batch and are not in the training device.  Generally speaking, designing a back-propagation rule for VQ under the mini-batch training setup is a challenging new problem.

\begin{figure}[t]
\begin{floatrow}
\ffigbox{%
\adjustbox{max width=.31\textwidth}{%
\includegraphics[width=\textwidth]{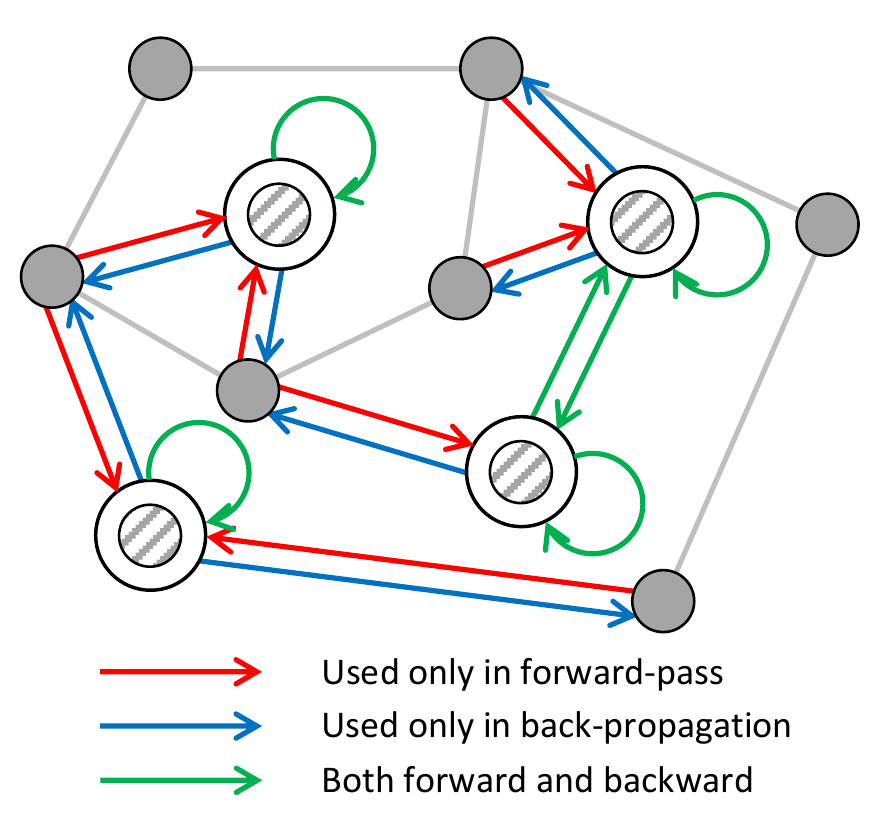}%
}}{%
{%
\caption{\label{fig:forward-backward} Three types of messages contribute to the mini-batch features and gradients. We only need ``red'' and ``green'' messages for the forward-pass. However, ``blue'' messages are required for back-propagation. The ``red'', ``blue'', and ``green'' messages are characterized by $\wt{C}\outof$, $(\wt{C^{\transs}})\outof$, and $C\within$ respectively (\cref{eq:approx-forward,eq:approx-backward}).}
\vspace{-5pt}
}%
}
\ffigbox{%
\adjustbox{max width=.50\textwidth}{%
\includegraphics[width=\textwidth]{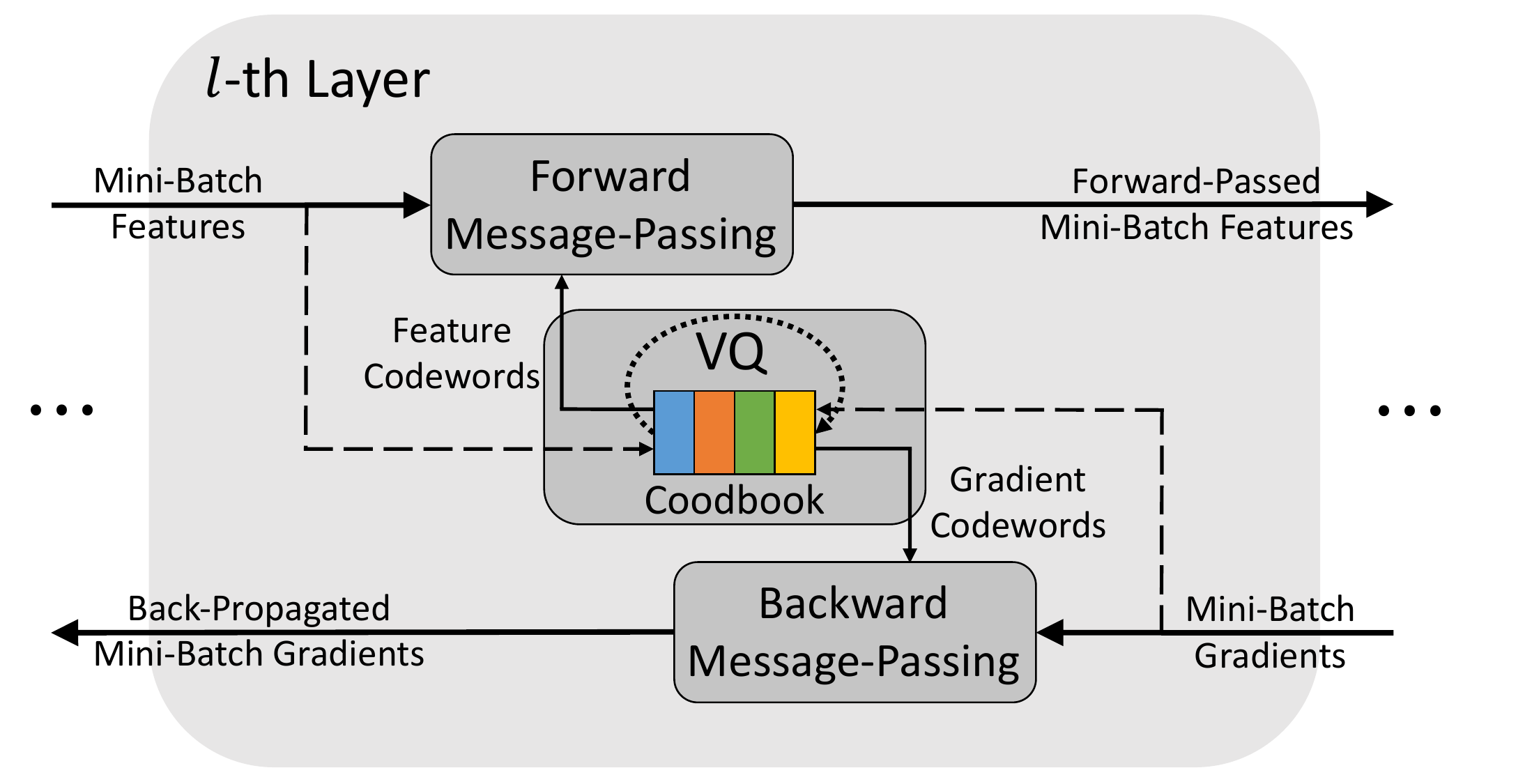}%
}}{%
{%
\caption{\label{fig:schema-arch}For each layer, VQ-GNN estimates the forward-passed mini-batch features using the previous layer's mini-batch features and the feature codewords through approximated forward message-passing (\cref{eq:approx-forward}). The back-propagated mini-batch gradients are estimated in a symmetric manner with the help of gradient codewords (\cref{eq:approx-backward}).}
\vspace{-5pt}
}%
}
\end{floatrow}
\vspace{-5pt}
\end{figure}

It is helpful to re-examine what is happening when we back-propagate on the full graph. In~\cref{sec:background}, we see that back-propagation of a layer of convolution-based GNN can also be realized by message passing (\cref{eq:gnn-backward}). In~\cref{fig:forward-backward}, we show the messages related to a mini-batch of nodes can be classified into three types. The ``green'' and ``red'' messages are the intra-mini-batch messages and the messages from out-of-mini-batch nodes, respectively. Apart from them, although the ``blue'' messages to out-of-mini-batch nodes do not contribute to the forward-passed mini-batch features, they are used during back-propagation and are an important part of the back-propagated mini-batch gradients. Since both forward-pass and back-propagation can be realized by message passing, can we approximate the back-propagated mini-batch gradients $\grad{X^{(l)}_B}$ in a symmetric manner? We can introduce a set of gradient codewords $\wt{G}^{(l+1)}=\diaginv{R\tran\vone_n}R\tran G^{(l+1)}$ using the same assignment matrix, where $G^{(l+1)} = \grad{\apx{X}^{(l+1)}}\odot\nonlinear'\big(\nonlinear^{-1}(X^{(l+1)})\big)$ is the gradients back-propagated through non-linearity. Each gradient codeword corresponds one-to-one with a feature codeword since we want to use only one assignment matrix $R$. Each pair of codewords are concatenated together during VQ updates. Following this idea, we define the approximated forward and backward message passing as follows:
\begin{equation}
\small
\label{eq:approx-forward}
    \begin{bmatrix}[1.4] \apx{X}^{(l+1)}_B \\ \bigcdot \end{bmatrix} = \nonlinear\Bigg(\sum_{s} \underbrace{\begin{bmatrix}[1.4] C^{(l,s)}\within & \wt{C}^{(l,s)}\outof \\ (\wt{C^{(l,s)\transs}})\outof & \bm{0} \end{bmatrix}}_{\substack{\text{approx. message passing} \\ \text{weight matrix }\aC^{(l,s)}}} \underbrace{\begin{bmatrix}[1.4] X^{(l)}_B \\ \wt{X}^{(l)} \end{bmatrix}}_{\substack{\text{mini-batch features} \\ \text{and feat. codewords}}} W^{(l,s)}\Bigg),
\end{equation}
\vspace{-5pt}
\begin{equation}
\small
\label{eq:approx-backward}
    \begin{bmatrix}[1.4] \apxgrad{X^{(l)}_B} \\ \bigcdot \end{bmatrix} = \sum_{s}\left(\aC^{(l,s)}\right)\tran \underbrace{\begin{bmatrix}[1.4] G^{(l+1)}_B \\ \wt{G}^{(l+1)} \end{bmatrix}}_{\substack{\text{mini-batch gradients} \\ \text{and grad. codewords}}} \left(W^{(l,s)}\right)\tran,
\end{equation}
where $\aC^{(l,s)}\in\sR^{(b+m)\times(b+m)}$ is the approximated message passing weight matrix and is shared during the forward-pass and back-propagation process. The lower halves of the left-hand side vectors of~\cref{eq:approx-forward,eq:approx-backward} are used in neither the forward nor the backward calculations and are never calculated during training or inference. The approximated forward and backward message passing enables the end-to-end mini-batch training and inference of GNNs and is the core of our VQ-GNN framework.


\cm{Error-bounds on estimated features and gradients.}
We can effectively upper bound the estimation errors of mini-batch features and gradients using the relative error $\epsilon$ of VQ under some mild conditions. For ease of presentation, we assume the GNN has only one convolution matrix in the following theorems.
\begin{theorem}
\label{thm:feat-error}
If the VQ relative error of $l$-th layer is $\epsilon^{(l)}$, the convolution matrix $C^{(l)}$ is either fixed or learnable with the Lipschitz constant of $h_{\theta^{(l)}}(\cdot):\sR^{2f_l}\to\sR$ upper-bounded by $\lip(h_{\theta^{(l)}})$, and the Lipschitz constant of the non-linearity is $\lip(\nonlinear)$, then the estimation error of forward-passed mini-batch features satisfies,
\begin{equation}
\label{eq:thm:feat-error}
    \|\apx{X}_B^{(l+1)}-X_B^{(l+1)}\|_F\leq \epsilon^{(l)} \cdot (1+O(\lip(h_{\theta^{(l)}})))\lip(\nonlinear)\|C^{(l)}\|_F\|X^{(l)}\|_F\|W^{(l)}\|_F.
\end{equation}
\end{theorem}

\begin{corollary}
\label{coro:grad-error}
If the conditions in~\cref{thm:feat-error} hold and the non-linearity satisfies $|\nonlinear'(z)|\leq\nonlinear'_{\max}$ for any $z\in\sR$, then the estimation error of back-propagated mini-batch gradients satisfies,
\begin{equation}
\label{eq:thm:grad-error}
    \|\apxgrad{X_B^{(l)}}-\grad{X_B^{(l)}}\|_F\leq \epsilon^{(l)} \cdot (1+O(\lip(h_{\theta^{(l)}}))\nonlinear'_{\max}\|C^{(l)}\|_F\|\grad{X^{(l+1)}}\|_F\|W^{(l)}\|_F.
\end{equation}
\end{corollary}
Note that the error bounds rely on the Lipschitz constant of $h(\cdot)$ when the convolution matrix is learnable. In practice, we can Lipshitz regularize GNNs like GAT~\citep{velivckovic2017graph} without affecting their performance; see~\cref{apd:algorithm}.


\cm{VQ-GNN: the complete algorithm and analysis of scalability.}
The only remaining question is how to update the learned codewords and assignments during training? In this paper, we use the VQ update rule proposed in~\citep{oord2017neural}, which updates the codewords as exponential moving averages of the mSeini-batch inputs; see~\cref{apd:algorithm} for the detailed algorithm. We find such an exponential moving average technique suits us well for the mini-batch training of GNNs and resembles the \emph{online k-means} algorithm. See~\cref{fig:schema-arch} for the schematic diagram of VQ-GNN, and the complete pseudo-code is in~\cref{apd:algorithm}. 

With VQ-GNN, we can mini-batch train and perform inference on large graphs using GNNs, just like a regular neural network (e.g., MLP). We have to maintain a small codebook of $k$ codewords and update it for each iteration, which takes an extra $O(Lkf)$ memory and $O(Lnkf)$ training time per epoch, where $L$ and $f$ are the numbers of layers and (hidden) features of the GNN respectively. We can effectively preserve all messages related to a mini-batch while randomly sampling nodes from the graph. The number of intra-mini-batch messages is $O(b^2d/n)$ when the nodes are sampled randomly. Thus we only need to pass $O(b^2d/n+bk)$ messages per iteration and $O(bd+nk)$ per epoch. In practice, when combined with techniques including product VQ  and implicit whitening (see~\cref{apd:algorithm}), we can further improve the stability and performance of VQ-GNN. These theoretical and experimental analyses justify the efficiency of the proposed VQ-GNN framework.

\section{Related Work}
\label{sec:related}

In this section, we review some of the recent scalable GNN methods and analyze their theoretical memory and time complexities, with a focus on scalable algorithms that can be universally applied to a variety of GNN models (like our VQ-GNN framework), including NS-SAGE\footnote{We call the neighbor sampling method in~\citep{hamilton2017inductive} NS-SAGE and the GNN model in the same paper SAGE-Mean to avoid ambiguity.}~\citep{hamilton2017inductive}, Cluster-GCN~\citep{chiang2019cluster}, and GraphSAINT~\citep{zeng2019graphsaint}. We consider GCN here as the simplest benchmark. For a GCN with $L$ layers and $f$-dimensional (hidden) features in each layer, when applied to a sparse graph with $n$ nodes and $m$ edges (i.e., average degree $d=m/n$) for ``full-graph'' training and inference: the memory usage is $O(Lnf+Lf^2)$ and the training/inference time is $O(Lmf+Lnf^2)$. We further assume the graph is large and consider the training and inference device memory is $O(b)$ where $b$ is the mini-batch size (i.e., the memory bottleneck limits the mini-batch size), and generally $d\ll b\ll n\ll m$ holds. We divide sampling baselines into three categories, and the complexities of selected methods are in~\cref{tab:complexities}. See~\cref{apd:related} for more related work discussions.


\begin{table}[t]
\centering
\caption{\label{tab:complexities}Memory and time complexities of sampling-based methods and our approach; see~\cref{sec:related} for details.}
\adjustbox{max width=1.0\textwidth}{%
\renewcommand*{\arraystretch}{1.5}
\begin{tabular}{Sl | Sc Sc Sc Sc} \toprule
Scalable Method  & Memory Usage                     & Pre-computation Time     & Training Time           & Inference Time          \\ \midrule
NS-SAGE                 & $O(br^Lf+Lf^2)$           & ---                      & $O(nr^L f+nr^{L-1}f^2)$ & \multirow{3}{*}{\begin{tabular}[c]{@{}c@{}}$\big|$\\$O(nd^L f+nd^{L-1}f^2)$\\$\big|$\end{tabular}} \\
Cluster-GCN             & $O(Lbf+Lf^2)$             & $O(m)$                   & $O(Lmf+Lnf^2)$          &                                                                                                          \\
GraphSAINT-RW           & $O(L^2bf+Lf^2)$           & ---                      & $O(L^2nf+L^2nf^2)$      &                                                                                                          \\ \arrayrulecolor{black!50} \midrule
\textbf{VQ-GNN (Ours)}  & $O(Lbf+Lf^2+Lkf)$          & ---                      & $O(Lbdf+Lnf^2+Lnkf)$    & $O(Lbdf+Lnf^2)$   \\ \bottomrule         
\end{tabular}}
\vspace{-5pt}
\end{table}


\cm{Neighbor-sampling.} Neighbor sampling scheme chooses a subset of neighbors in each layer to reduce the amount of message passing required. NS-SAGE~\citep{hamilton2017inductive} samples $r$ neighbors for each node and only aggregates the messages from the sampled node. For a GNN with $L$ layers, $O(br^L)$ nodes are sampled in a mini-batch, which leads to the complexities growing exponentially with the number of layers $L$; see~\cref{tab:complexities}. Therefore, NS-SAGE is not scalable on large graphs for a model with an arbitrary number of layers. NS-SAGE requires all the neighbors to draw non-stochastic predictions in the inference phase, resulting in a $O(d^L)$ inference time since we cannot fit $O(n)$ nodes all at once to the device. VR-GCN~\citep{chen2018stochastic} proposes a variance reduction technique to further reduce the size $r$ of sampled neighbors. However, VR-GCN requires a $O(Lnf)$ side memory of all the nodes' hidden features and suffers from this added memory complexity.


\cm{Layer-sampling.} These methods perform node sampling independently in each layer, which results in a constant sample size across all layers and limits the exponential expansion of neighbor size. FastGCN~\citep{chen2018fastgcn} applies importance sampling to reduce variance. Adapt~\citep{huang2018adaptive} improves FastGCN by an additional sampling network but also incurs the significant overhead of the sampling algorithm.


\cm{Subgraph-sampling.} Some proposed schemes sample a subgraph for each mini-batch and perform forward and backward passes on the same subgraph across all layers. Cluster-GCN~\citep{chiang2019cluster} partitions a large graph into several densely connected subgraphs and samples a subset of subgraphs (with edges between clusters added back) to train in each mini-batch. Cluster-GCN requires $O(m)$ pre-computation time and $O(bd)$ time to recover the intra-cluster edges when loading each mini-batch. GraphSAINT~\citep{zeng2019graphsaint} samples a set of nodes and takes the induced subgraph for mini-batch training. We consider the best-performing variant, GraphSAINT-RW, which uses $L$ steps of random walk to induce subgraph from $b$ randomly sampled nodes. $O(Lb)$ nodes and edges are covered in each of the $n/b$ mini-batches. Although $O(Ln)$ nodes are sampled with some repetition in an epoch, the number of edges covered (i.e., messages considered in each layer of a GNN) is also $O(Ln)$ and is usually much smaller than $m$. GraphSAINT-Node, which randomly samples nodes for each mini-batch, does not suffer from this $L$ factor in the complexities. However, its performance is worse than GraphSAINT-RW's. Like NS-SAGE and some other sampling methods, Cluster-GCN and GraphSAINT-RW cannot draw predictions on a randomly sampled subgraph in the inference phase. Thus they suffer from the same $O(d^L)$ inference time complexity as NS-SAGE; see~\cref{tab:complexities}.

\section{Experiments}
\label{sec:experiments}

In this section, we verify the efficiency, robustness, and universality of VQ-GNN using a series of experiments. See~\cref{apd:implementation} for implementation details and~\cref{apd:more-experiments} for ablation studies and more experiments.


\begin{table}[t]
\centering
\caption{\label{tab:efficiency-memory}Peak memory usage. Evaluated when fixing the number of gradient-descended nodes or the number of messages passed per mini-batch to be the same. Tested for GCN and SAGE-Mean on the \textit{ogbn-arxiv} benchmark.}
\adjustbox{max width=.65\textwidth}{%
\renewcommand*{\arraystretch}{1.2}
\begin{tabular}{Sl| Sc Sc | Sc Sc} \toprule
Fixed                   & \multicolumn{2}{c|}{\begin{tabular}[c]{@{}c@{}}  85K nodes per batch \end{tabular}} & \multicolumn{2}{c}{\begin{tabular}[c]{@{}c@{}} 1.5M messages passed per batch\end{tabular}} \\
GNN Model               & GCN                & SAGE-Mean                & GCN                & SAGE-Mean                \\ \midrule
NS-SAGE                 & ---                & \siunit{1140.3}{MB}             & ---                 &  \siunit{953.7}{MB} \\
Cluster-GCN             & \siunit{501.5}{MB}   & \siunit{514.1}{MB}              &   \siunit{757.4}{MB}  &  \siunit{769.3}{MB}  \\
GraphSAINT-RW           & \siunit{526.5}{MB}   & \siunit{519.2}{MB}              &   \siunit{661.6}{MB}  &  \siunit{650.4}{MB}  \\ \arrayrulecolor{black!50} \midrule
\textbf{VQ-GNN (Ours)}  & \siunit{758.0}{MB}   & \siunit{801.8}{MB}              &   \siunit{485.5}{MB}  & \siunit{508.5}{MB}    \\ \bottomrule
\end{tabular}}
\vspace{-5pt}
\end{table}


\cm{Scalability and efficiency: memory usage, convergence, training and inference time.}
We summarize the \emph{memory usage} of scalable methods and our VQ-GNN framework in~\cref{tab:efficiency-memory}. Based on the implementations of the PyG library~\cite{fey2019fast}, memory consumption of GNN models usually grows linearly with respect to both the number of nodes and the number of edges in a mini-batch. On the \textit{ogbn-arxiv} benchmark, we fix the number of gradient-descended nodes and the number of messages passed in a mini-batch to be $85$K and $1.5$M respectively for fair comparisons among the sampling methods and our approach. VQ-GNN might require some small extra memory when provided with the same amount of nodes per batch, which is the cost to retain all the edges from the original graph. However, our VQ-GNN framework can effectively preserve all the edges connected to a mini-batch of nodes (i.e., never drop edges); see~\cref{fig:schema-graph}. Thus when we fix the number of messages passed per batch, our method can show significant memory efficiency compared with the sampling baselines.

\begin{figure}[t]
    \centering
    \begin{subfigure}[b]{0.48\textwidth}
        \centering
        \includegraphics[trim = 0pt 0pt 0pt 0pt, clip, width=\textwidth]{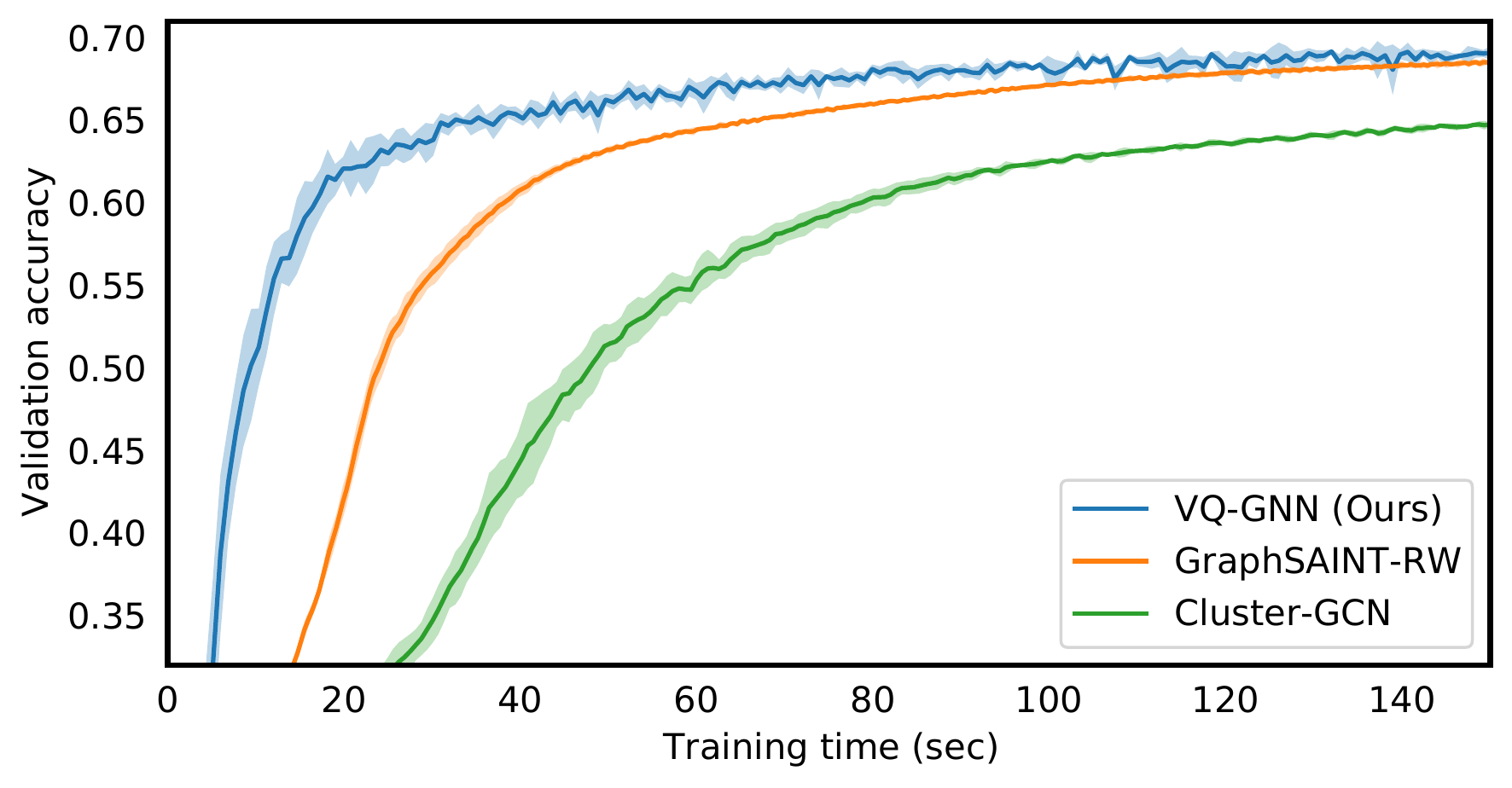}
        \caption{GCN}
        \label{fig:efficiency-convergence-gcn}
    \end{subfigure}
    \hfill
    \begin{subfigure}[b]{0.48\textwidth}
        \centering
        \includegraphics[trim = 0pt 0pt 0pt 0pt, clip, width=\textwidth]{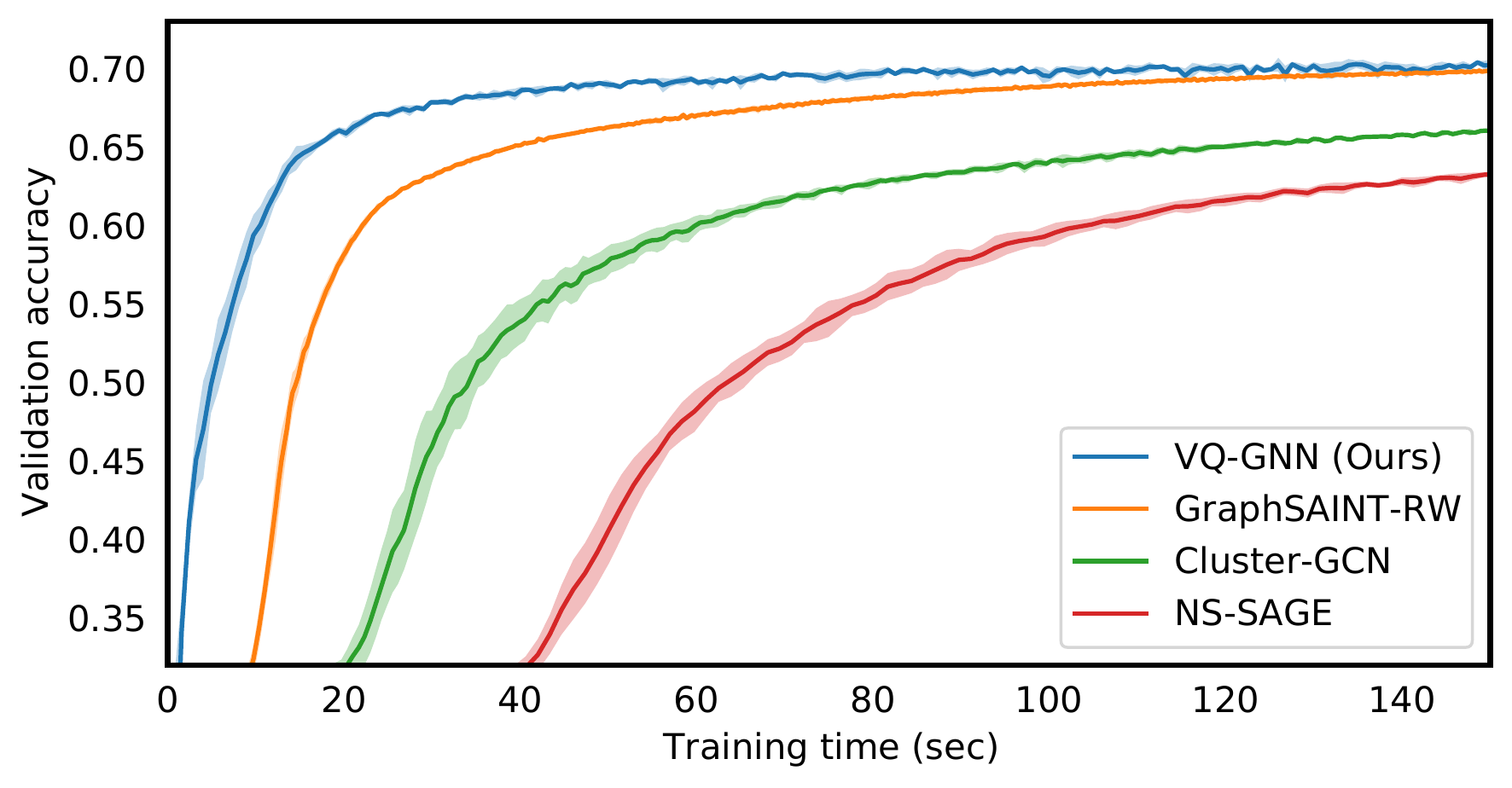}
        \caption{SAGE-Mean}
        \label{fig:efficiency-convergence-sage}
    \end{subfigure}
    \vspace{-5pt}
    \caption{\label{fig:efficiency-converge}Convergence curves (validation accuracy vs. training time). Mini-batch size and learning rate are kept the same. Tested for GCN and SAGE-Mean on the \textit{ogbn-arxiv} benchmark.}
\end{figure}

\cref{fig:efficiency-converge} shows the convergence comparison of various scalability methods, where we see VQ-GNN is superior in terms of the \emph{convergence speed} with respect to the \emph{training time}. When training GCN and SAGE-Mean on the \textit{ogbn-arxiv} benchmark for a specific amount of time (e.g., \siunit{100}{s}), the validation performance of VQ-GNN is always the highest. The training time in~\cref{fig:efficiency-converge} excludes the time for data loading, pre-processing, and validation set evaluation.

Our VQ-GNN approach also leads to compelling \emph{inference} speed-ups. Despite the training-efficiency issues of GNNs, conducting inference on large-scale graphs suffers from some unique challenges. According to our discussions in~\cref{sec:related}, and following the standard implementations provided by the Open Graph Benchmark (OGB)~\citep{hu2020open}, the three sampling-based baselines (which share the same inference procedure) require all of the $L$-hop neighbors of the mini-batch nodes to lie on the device at once during the inference phase. The inference time of SAGE-Mean trained with sampling-methods on the \textit{ogbn-arxiv} benchmark is \siunit{1.61}{s}, while our method can accelerate inference by an order of magnitude and reduce the inference time to \siunit{0.40}{s}.

\begin{table}[t]
\centering
\caption{\label{tab:performance}Performance comparison between sampling-based baselines and our approach, VQ-GNN.}
\adjustbox{max width=\textwidth}{%
\begin{threeparttable}
\begin{tabular}{Sl| Sc Sc Sc Sc Sc Sc} \toprule
\multicolumn{1}{c|}{\begin{tabular}[c]{@{}c@{}}Task\\ Benchmark \end{tabular}} & \multicolumn{3}{c}{\begin{tabular}[c]{@{}c@{}}Node Classification (Transductive)\\ \textit{ogbn-arxiv} (Acc.$\pm$std.) \end{tabular}} & \multicolumn{3}{c}{\begin{tabular}[c]{@{}c@{}}Node Classification (Transductive)\\ \textit{Reddit} (Acc.$\pm$std.) \end{tabular}} \\ \arrayrulecolor{black!50} \midrule
GNN Model               & GCN & SAGE-Mean & GAT & GCN & SAGE-Mean & GAT \\ \midrule
``Full-Graph''          & $.7029\pm.0036$ & $.6982\pm.0038$ & $.7097\pm.0035$ & OOM\tnote{2} & OOM\tnote{2} & OOM\tnote{2} \\ \arrayrulecolor{black!50} \midrule
NS-SAGE.                & NA\tnote{1} & $.7094\pm.0060$ & $.7123\pm.0044$ & NA\tnote{1} & $.9615\pm.0089$ & $.9426\pm.0043$ \\
Cluster-GCN             & $.6805\pm.0074$ & $.6976\pm.0049$ & $.6960\pm.0062$ & $.9264\pm.0034$ & $.9456\pm.0061$ & $.9380\pm.0055$ \\
GraphSAINT-RW           & $.7079\pm.0057$ & $.6987\pm.0039$ & $.7117\pm.0032$ & $.9225\pm.0057$ & $.9581\pm.0074$ & $.9431\pm.0067$ \\ \arrayrulecolor{black!50} \midrule
\textbf{VQ-GNN (Ours)}  & $.7055\pm.0033$ & $.7028\pm.0047$ & $.7043\pm.0034$ & $.9399\pm.0021$ & $.9449\pm.0024$ & $.9438\pm.0059$ \\ \arrayrulecolor{black} \bottomrule
\end{tabular}
\vspace{5pt}
\begin{tabular}{Sl| Sc Sc Sc Sc Sc Sc} \toprule
\multicolumn{1}{c|}{\begin{tabular}[c]{@{}c@{}}Task\\ Benchmark \end{tabular}} & \multicolumn{3}{c}{\begin{tabular}[c]{@{}c@{}}Node Classification (Inductive)\\ \textit{PPI} (F1-score\tnote{3} $\pm$std.) \end{tabular}} & \multicolumn{3}{c}{\begin{tabular}[c]{@{}c@{}}Link Prediction (Transductive)\\ \textit{ogbl-collab} (Hits@50$\pm$std.) \end{tabular}} \\ \arrayrulecolor{black!50} \midrule
GNN Model               & GCN & SAGE-Mean & GAT & GCN & SAGE-Mean & GAT \\ \midrule
``Full-Graph''          & $.9173\pm.0039$ & $.9358\pm.0046$ & $.9722\pm.0035$ & $.4475\pm.0107$ & $.4810\pm.0081$  & $.4048\pm.0125$ \\ \arrayrulecolor{black!50} \midrule
NS-SAGE.                & NA\tnote{1} & $.9121\pm.0033$ & $.9407\pm.0025$ & NA\tnote{1} & $.4776\pm.0041$ & $.3499\pm.0142$ \\
Cluster-GCN             & $.8852\pm.0066$ & $.8810\pm.0091$ & $.9051\pm.0077$ & $.4068\pm.0096$ & $.3486\pm.0216$ & $.3905\pm.0152$ \\
GraphSAINT-RW           & $.9110\pm.0057$ & $.9382\pm.0074$ & $.9612\pm.0042$ & $.4368\pm.0169$ & $.3359\pm.0128$ & $.3489\pm.0114$ \\ \arrayrulecolor{black!50} \midrule
\textbf{VQ-GNN (Ours)}  & $.9549\pm.0058$ & $.9578\pm.0019$ & $.9737\pm.0033$ & $.4316\pm.0134$ & $.4673\pm.0164$ & $.4102\pm.0099$ \\ \arrayrulecolor{black} \bottomrule
\end{tabular}
\appto\TPTnoteSettings{\normalsize}
\begin{tablenotes}[para]
    \item[1] NS-SAGE sampling method is not compatible with the GCN backbone.
    \item[2] ``OOM'' refers to ``out of memory''. The ``full-graph'' training on the \textit{Reddit} benchmark requires more than \siunit{11}{GB} of memory.
    \item[3] The \textit{PPI} benchmark comes with multiple labels per node, and the evaluation metric is F1 score instead of accuracy.
\end{tablenotes}
\end{threeparttable}}
\vspace{-5pt}
\end{table}

\cm{Performance comparison across various datasets, settings, and tasks.}
We validate the efficacy of our method on various benchmarks in~\cref{tab:performance}. The four representative benchmarks are selected because they have very different types of datasets, settings, and tasks. The \textit{ogbn-arxiv} benchmark is a common citation network of arXiv papers, while \textit{Reddit} is a very dense social network of Reddit posts, which has much more features per node and larger average node degree; see~\cref{tab:dataset-stats} in~\cref{apd:implementation} for detailed statistics of datasets. \textit{PPI} is a node classification benchmark under the inductive learning setting, i.e., neither attributes nor connections of test nodes are present during training, while the other benchmarks are all transductive. VQ-GNN can be applied under the inductive setting with only one extra step: during the inference stage, we now need to find the codeword assignments (i.e., the nearest codeword) of the test nodes before making predictions since we have no access to the test nodes during training. Neither the learned codewords nor the GNN parameters are updated during inference. \textit{ogbl-collab} is a link prediction benchmark where the labels and loss are intrinsically different.

It is very challenging for a scalable method to perform well on all benchmarks. In~\cref{tab:performance}, we confirm that VQ-GNN is more \emph{robust} than the three sampling-based methods. Across the four benchmarks, VQ-GNN can always achieve performance similar with or better than the oracle ``full-graph'' training performance, while the other scalable algorithms may suffer from performance drop in some cases. For example, NS-SAGE fails when training GAT on \textit{ogbl-collab}, Cluster-GCN consistently falls behind on \textit{PPI}, and GraphSAINT-RW's performance drops on the \textit{ogbl-collab} when using SAGE-Mean and GAT backbones. We think the robust performance of VQ-GNN is its unique value among the many other scalable solutions. VQ-GNN framework is robust because it provides bounded approximations of ``full-graph'' training (\cref{thm:feat-error,coro:grad-error}), while most of the other scalable algorithms do not enjoy such a theoretical guarantee. VQ-GNN is also \emph{universal} to various backbone models, including but not limited to GCN, SAGE-Mean, and GAT shown here; see~\cref{apd:more-experiments} for more experiments on GNNs that utilize multi-hop neighborhoods and global context, e.g., graph transformers.

\section{Conclusion}
\label{sec:conclusion}


\cm{Summary of our framework: strengths, weaknesses, future directions, and broader impacts.}
This paper introduced the proposed VQ-GNN framework, which can scale most state-of-the-art GNNs to large graphs through a principled and fundamentally different approach compared with sampling-based methods. We have shown both theoretically and experimentally that our approach is efficient in memory usage, training and inference time, and convergence speed. VQ-GNN can be universally applied to most GNN models and different graph learning tasks and can equivalently scale-up GNNs utilizing many-hops-away or global context for each layer. However, the performance of VQ-GNN relies on the quality of approximation provided by VQ. In practice, for VQ to work adequately in GNN, a set of techniques are necessary. Because of the limited time, we did not heuristically explore all possible techniques or optimize the VQ design. Given that our preliminary design of VQ in GNN already achieved competitive performance compared with the state-of-the-art sampling baselines, we hypothesize that further optimization of VQ design could improve performance. We hope our work opens up promising new avenues of research for scaling up GNNs, which also has the potential to be applied to other data domains wherever the size of a single sample is large, e.g., long time-series or videos. Considering broader impacts, we view our work mainly as a methodological and theoretical contribution, which paves the way for more resource-efficient graph representation learning. We envision our methodological innovations can enable more scalable ways to do large-network analysis for social good. However, progress in graph embedding learning might also trigger other hostile social network analyses, e.g., extracting fine-grained user interactions for social tracking.



\begin{ack}

Goldstein, Kong, and Chen were supported by the Office of Naval Research, AFOSR MURI program, the DARPA Young Faculty Award, and the National Science Foundation Division of Mathematical Sciences. Additional support was provided by Capital One Bank and JP Morgan Chase. Huang and Ding were supported by a startup fund from the Department of Computer Science of the University of Maryland, National Science Foundation IIS-1850220 CRII Award 030742-00001, DOD-DARPA-Defense Advanced Research Projects Agency Guaranteeing AI Robustness against Deception (GARD), Air Force Material Command, and Adobe, Capital One and JP Morgan faculty fellowships. Li and Dickerson were supported in part by NSF CAREER Award IIS-1846237, NSF D-ISN Award \#2039862, NSF Award CCF-1852352, NIH R01 Award NLM-013039-01, NIST MSE Award \#20126334, DARPA GARD \#HR00112020007, DoD WHS Award \#HQ003420F0035, ARPA-E Award \#4334192 and a Google Faculty Research Award.


\end{ack}


{
\small
\bibliographystyle{unsrtnat}
\typeout{}
\bibliography{reference}
}


\section*{Checklist}

The checklist follows the references.  Please
read the checklist guidelines carefully for information on how to answer these
questions.  For each question, change the default \answerTODO{} to \answerYes{},
\answerNo{}, or \answerNA{}.  You are strongly encouraged to include a {\bf
justification to your answer}, either by referencing the appropriate section of
your paper or providing a brief inline description.  For example:
\begin{itemize}
  \item Did you include the license to the code and datasets? \answerYes{See Section.}
  \item Did you include the license to the code and datasets? \answerNo{The code and the data are proprietary.}
  \item Did you include the license to the code and datasets? \answerNA{}
\end{itemize}
Please do not modify the questions and only use the provided macros for your
answers.  Note that the Checklist section does not count towards the page
limit.  In your paper, please delete this instructions block and only keep the
Checklist section heading above along with the questions/answers below.

\begin{enumerate}

\item For all authors...
\begin{enumerate}
  \item Do the main claims made in the abstract and introduction accurately reflect the paper's contributions and scope?
    \answerYes{}
  \item Did you describe the limitations of your work?
    \answerYes{} See~\cref{sec:conclusion}.
  \item Did you discuss any potential negative societal impacts of your work?
    \answerYes{} See~\cref{sec:conclusion}.
  \item Have you read the ethics review guidelines and ensured that your paper conforms to them?
    \answerYes{}
\end{enumerate}

\item If you are including theoretical results...
\begin{enumerate}
  \item Did you state the full set of assumptions of all theoretical results?
    \answerYes{} See~\cref{thm:jl-lemma,thm:feat-error,coro:grad-error}.
	\item Did you include complete proofs of all theoretical results?
    \answerYes{} See~\cref{apd:proofs}.
\end{enumerate}

\item If you ran experiments...
\begin{enumerate}
  \item Did you include the code, data, and instructions needed to reproduce the main experimental results (either in the supplemental material or as a URL)?
    \answerYes{}
  \item Did you specify all the training details (e.g., data splits, hyperparameters, how they were chosen)?
    \answerYes{} See~\cref{apd:implementation} and the code.
	\item Did you report error bars (e.g., with respect to the random seed after running experiments multiple times)?
    \answerYes{}
	\item Did you include the total amount of compute and the type of resources used (e.g., type of GPUs, internal cluster, or cloud provider)?
    \answerYes{} See~\cref{apd:implementation}.
\end{enumerate}

\item If you are using existing assets (e.g., code, data, models) or curating/releasing new assets...
\begin{enumerate}
  \item If your work uses existing assets, did you cite the creators?
    \answerYes{}
  \item Did you mention the license of the assets?
    \answerYes{}
  \item Did you include any new assets either in the supplemental material or as a URL?
    \answerNo{}
  \item Did you discuss whether and how consent was obtained from people whose data you're using/curating?
    \answerNA{}
  \item Did you discuss whether the data you are using/curating contains personally identifiable information or offensive content?
    \answerNA{}
\end{enumerate}

\item If you used crowdsourcing or conducted research with human subjects...
\begin{enumerate}
  \item Did you include the full text of instructions given to participants and screenshots, if applicable?
    \answerNA{}
  \item Did you describe any potential participant risks, with links to Institutional Review Board (IRB) approvals, if applicable?
    \answerNA{}
  \item Did you include the estimated hourly wage paid to participants and the total amount spent on participant compensation?
    \answerNA{}
\end{enumerate}

\end{enumerate}


\clearpage
\newpage
{\centering \bf \LARGE
    Supplementary Material
}
\appendix


\section{Generalized Graph Convolution Framework}
\label{apd:generalized-conv}
In this section, we present more results and discussions regarding the common generalized graph convolution framework in~\cref{sec:background}.

\cm{Summary of GNNs re-formulated into the common graph convolution framework.}
As stated in~\cref{sec:background}, most GNNs can be interpreted as performing message passing on node features, followed by feature transformation and an activation function (\cref{eq:approx-forward}), which is known as the common ``generalized graph convolution'' framework. We list more GNN models that fall into this framework in the following~\cref{tab:more-GNN-models}.

\begin{table}[ht]
\centering
\caption{\label{tab:more-GNN-models}Summary of more GNNs re-formulated as generalized graph convolution.}
\adjustbox{max width=\textwidth}{%
\begin{threeparttable}
\renewcommand*{\arraystretch}{1.5}
\begin{tabular}{Sc Sc Sc Sc Sl} \toprule
Model Name & Design Idea & Conv. Matrix Type & \# of Conv. & Convolution Matrix \\ \cmidrule{1-5}\morecmidrules\cmidrule{1-5}
GIN\tnote{1}~~\citep{xu2018powerful} & WL-Test & \multicolumn{1}{c}{\begin{tabular}[c]{@{}c@{}} Fixed $+$ \\ Learnable \end{tabular}} & $2$ & \multicolumn{1}{l}{$\left\{\begin{tabular}[c]{@{}l@{}} $C^{(1)}=A$\\ $\fC^{(2)}=I_n$\itand$h^{(2)}_{\epsilon^{(l)}}=1+\epsilon^{(l)}$ \end{tabular}\right.\kern-\nulldelimiterspace$} \\ \midrule
ChebNet\tnote{2}~~\citep{defferrard2016convolutional} & Spectral Conv. & Learnable & order of poly. & \multicolumn{1}{l}{$\left\{\begin{tabular}[c]{@{}l@{}} $\fC^{(1)}=I_n$, $\fC^{(2)}=2L/\lambda_{\max}-I_n$,~\\ $\fC^{(s)}=2\fC^{(2)}\fC^{(s-1)}-\fC^{(s-2)}$\\\itand$h^{(s)}_{\theta^{(s)}}=\theta^{(s)}$ \end{tabular}\right.\kern-\nulldelimiterspace$} \\
GDC\tnote{3}~~\citep{klicpera2019diffusion} & Diffusion & Fixed & $1$ & $C=S$ \\
Graph Transformers\tnote{4}~~\citep{yaronlipman2020global, rong2020self, zhang2020graph} & Self-Attention & Learnable & \# of heads & \multicolumn{1}{l}{$\left\{\begin{tabular}[c]{@{}l@{}} $\fC^{(s)}_{i,j}=1$\itand$h^{(s)}_{(W^{(l,s)}_Q, W^{(l,s)}_K)}(X^{(l)}_{i,:}, X^{(l)}_{j,:})$\\$=\exp\big(\frac{1}{\sqrt{d_{k,l}
}}(X^{(l)}_{i,:}W^{(l,s)}_Q)(X^{(l)}_{j,:}W^{(l,s)}_K)\tran\big)$  \end{tabular}\right.\kern-\nulldelimiterspace$} \\ \bottomrule
\end{tabular}
\appto\TPTnoteSettings{\large}
\begin{tablenotes}[para]
    \item[1] The weight matrices of the two convolution supports are the same, $W^{(l,1)}=W^{(l,2)}$.
    \item[2] Where normalized Laplacian $L=I_n-D^{-1/2}AD^{-1/2}$ and $\lambda_{\max}$ is its largest eigenvalue, which can be approximated as $2$ for a large graph.
    \item[3] Where $S$ is the diffusion matrix $S=\sum_{k=0}^\infty\theta_k\bm{T}^k$, for example, decaying weights $\theta_k=e^{-t}\frac{t^k}{k!}$ and transition matrix $\bm{T}=\wt{D}^{-1/2}\wt{A}\wt{D}^{-1/2}$.
    \item[4] Need row-wise normalization. Only describes the global self-attention layer, where $W^{(l,s)}_Q, W^{(l,s)}_Q\in\sR^{f_l, d_{k,l}}$ are weight matrices which compute the queries and keys vectors. In contrast to GAT, all entries of $\fC^{(l,s)}_{i,j}$ are non-zero. Different design of Graph Transformers~\citep{yaronlipman2020global, rong2020self, zhang2020graph} use graph adjacency information in different ways, and is not characterized here, see the original papers for details.
\end{tablenotes}
\end{threeparttable}}
\end{table}

\cm{Receptive fields.} The model in the top part of~\cref{tab:more-GNN-models} (GIN) and the models in~\cref{tab:GNN-models} in~\cref{sec:background} (GCN, SAGE-Mean, and GAT) follow direct-neighbor message passing, and their single-layer receptive fields, defined as a set of nodes $\gR^1_i$ whose features $\{X^{(l)}_{j,:}\mid j\in\gR_i\}$ determines $X^{(l+1)}_{i,:}$, are exactly $\gR^1_i=\{i\}\cup\gN_i$, where $\gN_i$ is the set of direct neighbors of node $i$. The models in the bottom part of~\cref{tab:more-GNN-models} (ChebNet, GDC, Transformer) can utilize many-hops-away or gloabl context each layer, and their single-layer receptive field $\gR^1_i\supseteq\{i\}\cup\gN_i$.

\cm{GNNs that cannot be defined as graph convolution.} Some GNNs, including Gated Graph Neural Networks~\citep{li2015gated} and ARMA Spectral Convolution Networks~\citep{bianchi2021graph} cannot be re-formulated into this common graph convolution framework because they rely on either Recurrent Neural Networks (RNNs) or some iterative processes, which are out of the paradigm of message passing. 

\cm{Further row-wise normalization of a learnable convolution matrix.} Sometimes a learnable convolution matrix many be further row-wise normalized as $C^{(l,s)}_{i,j} \gets C^{(l,s)}_{i,j}/\sum_j C^{(l,s)}_{i,j}$. This is required for self-attention-based GNNs, including GAT and Graph Transformers. This normalization process will not affect the single-layer receptive fields of the two models and can be handled by a small modification to the VQ-GNN algorithm; see~\cref{apd:algorithm}.


\section{Proofs of Theoretical Results}
\label{apd:proofs}
This section provides the formal proofs of the theoretical results in the paper.

\paragraph{Proof of \texorpdfstring{\cref{thm:jl-lemma}.}{Theorem \ref{thm:jl-lemma}}}
The proof is based on the sparse distributional Johnson–Lindenstrauss (JL) lemma in~\citep{kane2014sparser}, which is stated as the following~\cref{lemma:sparse-jl} for completeness.
\begin{lemma}
\label{lemma:sparse-jl}
For any integer $n>0$, and any $0<\varepsilon$, $\delta<1/2$, there exists a probability distribution of $R\in\sR^{n\times k}$ where $R$ has only an $O(\varepsilon)$-fraction of non-zero entries, such that for any $x\in\sR^d$,
\begin{equation}
\label{eq:lemma:sparse-jl}
    \Pr\left((1-\varepsilon)\|x\|_2\leq\|R\tran x\|_2\leq(1+\varepsilon)\|x\|_2\right) > 1-\delta',
\end{equation}
with $k=\Theta(\varepsilon^{-2}\log(1/\delta))$ and thus $\delta'=O(\exps{-\varepsilon^2k})$. 
\end{lemma}

\begin{prevproof}{thm:jl-lemma}
\label{proof:jl-lemma}
From the given lemma, consider the inner product $x\tran y$ between any two vectors $x, y\in\sR^n$. Using the union-bound, one can derive that,
\[
    \Pr\left(|x\tran R R\tran y - x\tran y| \leq \varepsilon|x\tran y|\right) > 1-2\delta'.
\]

Now, let $x$ be any row vector of $C$ and $y$ be any column vector of $X$, i.e., $x=C_{b,:}$ and $y=X_{:,a}$ for any $b\in\{1,\ldots,n\}$ and $a\in\{1,\ldots,f\}$, then we get,
\[
    \Pr\left(|C_{b,:} R R\tran X_{:,a} - C_{b,:} X_{:,a}| \leq \varepsilon|C_{b,:} X_{:,a}|\right) > 1-2\delta'.
\]

Therefore, by the union bound again, we have,
\[
\begin{split}
    &\Pr\left(\|C R R\tran X_{:,a}-CX_{:,a}\|_2 \leq \varepsilon\|CX_{:,a}\|_2\right) \\
    &\qquad \geq 1 - \sum_{b=1}^n \Pr\left(|C_{b,:} R R\tran X_{:,a} - C_{b,:} X_{:,a}| > \varepsilon|C_{b,:} X_{:,a}|\right) > 1 - 2n\delta'. \\
\end{split}
\]

Now we denote $\delta=2n\delta'$. The required $\delta=O(1/n)$ can be achieved when $k=\Theta(\log(n)/\varepsilon^2)$. Thus we finally obtain,
\[
    \Pr\left(\|C R R\tran X_{:,a}-CX_{:,a}\|_2<\varepsilon\|CX_{:,a}\|_2\right) > 1-\delta,
\]
with $k=\Theta(\log(n)/\varepsilon^2)$ and $\delta=O(1/n)$, which concludes the proof.
\end{prevproof}

\paragraph{Proof of \texorpdfstring{\cref{thm:feat-error}.}{Theorem \ref{thm:feat-error}}}
The proof is a direct application of the VQ relative-error bound and the Lipschitz continuity properties.

\begin{prevproof}{thm:feat-error}
\label{proof:feat-error}
If we denote the learnable convolution matrix calculated using the approximated features (i.e., feature codewords) by $C'^{(l)}$, then we have,
\[
\begin{split}
    \|\apx{X}_B^{(l+1)}-X_B^{(l+1)}\|_F & \leq \lip(\nonlinear)\|C'^{(l)} R~\diaginv{R\tran\vone_n} R\tran X^{(l)} W^{(l)} - C^{(l)} X^{(l)} W^{(l)}\|_F \\
    & \leq \lip(\nonlinear) \|C'^{(l)} R~\diaginv{R\tran\vone_n} R\tran X^{(l)} - C^{(l)} X^{(l)}\|_F \|W^{(l)}\|_F, \\
\end{split}
\]
where
\begin{equation}
\label{apd:eq:proof:feat-error}
\begin{split}
    & \|C'^{(l)} R~\diaginv{R\tran\vone_n} R\tran X^{(l)} - C^{(l)} X^{(l)}\|_F \\
    & \qquad \leq \|C'^{(l)}-C^{(l)}\|_F \|R~\diaginv{R\tran\vone_n} R\tran X^{(l)}\|_F \\
    & \qquad \qquad + \|C^{(l)}\|_F \|X^{(l)} - R~\diaginv{R\tran\vone_n} R\tran X^{(l)} \|_F. \\
\end{split}
\end{equation}

The relative error $\epsilon^{(l)}$ of VQ of the $l$-th layer is defined as,
\[
    \|X^{(l)} - R~\diaginv{R\tran\vone_n} R\tran X^{(l)} \|_F \leq \epsilon^{(l)} \|X^{(l)}\|_F.
\]
Thus the second term on the right-hand side of~\cref{apd:eq:proof:feat-error} satisfies,
\[
    \|C^{(l)}\|_F \|X^{(l)} - R~\diaginv{R\tran\vone_n} R\tran X^{(l)} \|_F \leq \epsilon^{(l)} \cdot \|C^{(l)}\|_F \|X^{(l)}\|_F
\]

For the first term on the right-hand side of~\cref{apd:eq:proof:feat-error} satisfies, first note that,
\[
    \|R~\diaginv{R\tran\vone_n} R\tran\|_F = \sqrt{k}
\]
is a constant where $k$ is the number of codewords.

And $C'^{(l)}$ is different to $C^{(l)}$, because the convolution matrix is learnable. We have
\[
    C^{(l)}_{i,j} = \fC_{i,j} h_{\theta^{(l)}}(X_{i,:}, X_{j,:}) \quad \text{and} \quad
    C'^{(l)}_{i,j} = \fC_{i,j} h_{\theta^{(l)}}(\wt{X}^{(l)}_{i,:}, \wt{X}^{(l)}_{j,:}),
\]
where $\wt{X}^{(l)}=\diaginv{R\tran\vone_n} R\tran X^{(l)}$. 

Therefore, using the Lipschitz constant of $h_{\theta^{(l)}}$, because for any $i\in\{1,\ldots,n\}$,
\[
    \|X_{i,:}- \wt{X}^{(l)}_{i,:}\|_2 \leq \|X^{(l)} - R~\diaginv{R\tran\vone_n} R\tran X^{(l)} \|_F \leq \epsilon^{(l)} \|X^{(l)}\|_F,
\]
we have for any $i, j\in\{1,\ldots,n\}$,
\[
    |C'^{(l)}_{i,j}-C^{(l)}_{i,j}| \leq 2|\fC_{i,j}| \cdot \lip(h_{\theta^{(l)}}) \epsilon^{(l)} \|X^{(l)}\|_F
\]
Summing up for all $(i,j)\in\{1,\ldots,n\}^2$, we have, for the first term on the right-hand side of~\cref{apd:eq:proof:feat-error},
\[
    \|C'^{(l)}-C^{(l)}\|_F \leq 2\|\fC\|_F \cdot \lip(h_{\theta^{(l)}}) \epsilon^{(l)} \|X^{(l)}\|_F
\]

Combining these two inequalities, we finally have,
\[
    \|\apx{X}_B^{(l+1)}-X_B^{(l+1)}\|_F\leq \epsilon^{(l)} \cdot (1+O(\lip(h_{\theta^{(l)}})))\lip(\nonlinear)\|C^{(l)}\|_F\|X^{(l)}\|_F\|W^{(l)}\|_F.
\]
which concludes the proof.
\end{prevproof}

\cm{Proof of \texorpdfstring{\cref{coro:grad-error}}{Corollary \ref{coro:grad-error}}}
The proof is similar to the proof of \cref{thm:feat-error}.

\begin{prevproof}{coro:grad-error}
\label{proof:grad-error}
This time, we use the message passing equation for the back-propagation process, \cref{eq:gnn-backward}, restated as follows,
\[
    \grad{X^{(l)}} = \sum_{s}\left(C^{(l,s)}\right)\tran\bigg(\grad{X^{(l+1)}}\odot\nonlinear'\Big(\nonlinear^{-1}\big(X^{(l+1)}\big)\Big)\bigg)\left(W^{(l,s)}\right)\tran.
\]

Note that,
\[
    \|\grad{X^{(l+1)}}\odot\nonlinear'\Big(\nonlinear^{-1}\big(X^{(l+1)}\big)\Big)\|_F \leq \nonlinear'_{\max} \|\grad{X^{(l+1)}}\|_F,
\]
The rest of the proof simply follows the proof of~\cref{thm:feat-error}, we similarly obtain,
\[
    \|\apxgrad{X_B^{(l)}}-\grad{X_B^{(l)}}\|_F\leq \epsilon^{(l)} \cdot (1+O(\lip(h_{\theta^{(l)}}))\nonlinear'_{\max}\|C^{(l)}\|_F\|\grad{X^{(l+1)}}\|_F\|W^{(l)}\|_F,
\]
which concludes the proof.
\end{prevproof}


\section{More Theoretical Discussions}
\label{apd:theory}

\cm{Upper-bound the estimation error of learnable parameters' gradients.}
Given that the gradients of each node in each layer is approximated with bounded error (\cref{coro:grad-error}), it is not hard to see that the gradients of learnable parameters $W^{(l)}$ and $\theta^{(l)}$ are also estimated with bounded errors. Firstly, the gradients of $W^{(l)}$, $\grad{W^{(l)}}$, can be calculated from $\grad{X^{(l+1)}}$ before going into the approximated back-propagation process. Secondly, the gradients of $\theta^{(l)}$, $\grad{\theta^{(l)}}$ has bounded estimation error as long as for any $(i,j)\in\{1,\ldots,n\}^2$ that $|C'^{(l)}_{i,j}-C^{(l)}_{i,j}|$ is bounded, as we have shown in~\cref{proof:feat-error}.

\cm{Upper-bound the size of VQ codebook.}
In order to effectively upper-bound the size of the VQ codebook, i.e., the number of reference vectors in VQ and the number of clusters in \emph{k-means}, we generally need some extra mild assumptions on the distributions of the (hidden) features and gradients. For instance, if we assume the distributions of each feature and gradient in the $l$-th layer are \emph{sub-Gaussian}, i.e., they have strong tail decay dominated by the tails of a Gaussian. Then, we can show the relative error of VQ $\epsilon=O(k^{-1/f})$, where $f$ is the dimensionality of features and gradients. In addition to this, if we use product VQ (see~\cref{apd:algorithm}) to utilize multiple VQ process, each working on a subset of $f_\text{prod}\ll f$ features and gradients, then we can bound the size of codebook as $k=O(\epsilon^{-f_\text{prod}})$.

\cm{Justification of choosing VQ as the dimensionality reduction method.}
We choose VQ as the method of dimensionality reduction to scale up GNNs because of the following reasons:
\begin{itemize}[leftmargin=*, topsep=1.5pt]
\setlength\itemsep{0.75pt}
    \item Categorical constraint of VQ: We choose VQ mainly because of its categorical constraint, i.e., the rows of projection matrix $R$ are unit vectors, shown in Eq. (5) of the manuscript. Intuitively, this means each node feature vector corresponds to exactly one codeword vector at a time, and thus we can replace its feature vector with the corresponding codeword. This is what we mean by "node-identity-preserving" in~\cref{sec:insight,sec:method}.
    \item Compared with PCA: PCA is not suitable when we want to compress the feature table of $n$ nodes into a compact codebook when the number of nodes $n$ is much larger than the number of features per node. Moreover, PCA is shown to have the same objective function as VQ under some settings but without the categorical constraint~\citep{ding2004k}. In PCA, each node feature is represented using the set of principal components (i.e., eigenvectors of the covariance matrix). Thus, we must use the complete set of principal components to recover each node feature vector before passing it to GNNs. Moreover, it is much harder to develop an online PCA algorithm to be used along with the training of GNNs.
    \item Compared with Fisher's LDA: Fisher's LDA, compared with PCA, is supervised and takes class labels into consideration. However, node classification or link prediction on a large graph is a semi-supervised learning problem, and we do not have access to all the node labels during training. We do not know how to compress the test nodes' features under the transductive learning setting, and thus LDA is not a choice.
    \item Compared with randomized projection and locality-sensitive hashing: VQ is a better choice than randomized projection and locality-sensitive hashing because it is deterministic. Using VQ, we do not have to deal with the extra burden introduced by stochasticity.
\end{itemize}


\section{More Discussions of Related Work}
\label{apd:related}
In this section, we continue the discussions of related work in~\cref{sec:related} and review some other scalable or efficient methods for GNNs and beyond.

\cm{Other efficient methods for graph representation learning: MLP-based sample models and mixed-precision approaches.}
There exist some other MLP-based models, e.g., SGC~\citep{wu2019simplifying}, which requires only a one-time message passing during the pre-computation stage with $O(Lmf)$ time. However, SGC, PPRGo~\citep{bojchevski2020scaling} and SIGN~\citep{rossi2020sign} over-simplify the GNN model and limit the expressive power. Despite the fact that they are fast, their performance is not generally comparable with other GNNs. Degree-Quant~\citep{tailor2020degree} and SGQuant~\citep{feng2020sgquant} applies mixed-precision techniques specifically designed for GNNs to improve efficiency. Although Degree-Quant can reduce the runtime memory usage from 4x to 8x (SGQquant achieves 4.25-31.9x reduction), they did not provide a solution to effectively mini-batch the GNN training. We note that Degree-Quant and SGQuant do not provide means to mini-batch a large graph and are still ``full-graph'' training.

\cm{Outside graph learning: VQ in neural network and efficient self-attention.}
The general idea of using Vector Quantization (VQ) in a neural network is initially proposed for Variational Auto-Encoders (VAEs) in VQ-VAE~\citep{oord2017neural, razavi2019generating}, and is later generalized to other generative modeling~\citep{maaloe2019biva}, contrastive learning~\citep{caron2020unsupervised}, and self-supervised learning~\citep{baevski2019vq, baevski2020wav2vec}. Our work learns form their success and is one of the first attempts of applying VQ to large-scale semi-supervised learning. Our work also shares similarities with the recent advances of efficient self-attention techniques to speed up Transformer models. Linformer~\citep{wang2020linformer} linearizes the time and memory complexities of self-attention by projecting the inputs to a compact representation and approximating the self-attention scores by a low-rank matrix. Hamburger~\citep{geng2021attention} generally discussed the update rule and back-propagation problem of VQ.


\section{Algorithm Details}
\label{apd:algorithm}

\cm{The complete pseudo-code.}
Here, we present the complete pseudo-code of our VQ-GNN algorithm as~\cref{alg:vq-gnn}. Three important components of VQ-GNN, namely \emph{approximated forward message-passing}, \emph{approximated backward message-passing}, and \emph{VQ Update}, are highlighted in~\cref{alg:vq-gnn} and in~\cref{fig:schema-arch} in~\cref{sec:method}. 

\begin{algorithm}[ht]
\caption{\label{alg:vq-gnn}VQ-GNN: our proposed universal framework to scale most state-of-the-art Graph Neural Networks to large graphs using Vector Quantization. For ease of presentation, we assume the GNN has only one fixed convolution matrix.}
\small 
\begin{algorithmic}[1]
    \Require Input node features $X$, ground-truth labels $Y$
    \Require GNN's convolution matrix $C$ (fixed convolution) or $\fC$ and $h(\cdot,\cdot)$ (learnable convolution)
    \Procedure{VQ-GNN$_{C}$}{$X, Y$}
        \For{$l=0,\ldots,L-1$} \Comment{\textit{Initialization}}
            \parState{Randomly initialize GNN learnable parameters $W^{(l)}$ and $\theta^{(l)}$ and codewords $\wt{V}^{(l)}=\wt{X}^{(l)}\concat\wt{G}^{(l+1)}$ which are feature codewords $\wt{X}^{(l)}$ concatenated with gradient codewords $\wt{G}^{(l+1)}$}
            \State Initialize codeword assignment $R^{(l)}$ according to the initial codewords
        \EndFor
        
        \For{epoch $t=1,\ldots,T$}
            \For{indices $\seq{i_b}$ sampled from $\{1,\ldots,n\}$} \Comment{\textit{Mini-batch Training}}
                \parState{Load the mini-batch features $X_B$($=X_{\seq{i_b},:}$), labels $Y_B$, selected rows and columns of convolution matrix $C_B$ and $(C\tran)_B$, and selected rows of each codeword assignment matrix $R^{(l)}_B$ to the training device, and set $X^{(0)}_B \gets X_B$}
                
                \For{$l=0,\ldots,L-1$} \Comment{\textit{Forward-Pass}}
                    \parState{Compute the approximate message passing weight matrix $\aC^{(l)}$ using $C_B, C\tran_B, R^{(l)}_B$; see~\cref{eq:approx-forward}}
                    \parState{\func{Approximated Forward Message-Passing}: estimate next layer's features $X^{(l+1)}_B$ using previous layer's mini-batch features $X^{(l)}_B$ and feature codewords $\wt{X}^{(l)}$; see~\cref{eq:approx-forward}}
                \EndFor
                
                \parState{Compute $\ell=$ \Call{Loss}{$Y_B, \hat{Y}_B$}, where the predicted labels $\hat{Y}_B=$ \Call{Softmax}{$X^{(L)}_B$} (node classification) or \Call{LinkPred}{$X^{(L)}_B$} (link prediction), and set $G^{(L)}_B \gets \nabla_{X^{(L)}_B}\ell$ (since no non-linearity $\nonlinear$ in the last GNN layer; see~\cref{eq:approx-backward})}
                
                \For{$l=L-1,\ldots,0$} \Comment{\textit{Back-Propagation}}
                    \parState{\func{Approximated Backward Message-Passing}: estimate lower layer's gradients $G^{(l)}_B$ and $\grad{W^{(l)}}$ using higher layer's mini-batch gradients $G^{(l+1)}_B$ and gradient codewords $\wt{G}^{(l+1)}$; see~\cref{eq:approx-backward}}
                \EndFor
                
                \For{$l=0,\ldots,L-1$} \Comment{\textit{VQ Update}}
                    \parState{\func{VQ Update}: Update the concatenated codewords $\wt{V}^{(l)}=\wt{X}^{(l)}\concat\wt{G}^{(l+1)}$ and codeword assignment $R^{(l)}_B$ using this mini-batch's concatenated feature and gradient vectors $V^{(l)}=X^{(l)}_B\concat G^{(l+1)}_B$ and the old concatenated codewords} \label{alg:line:vq-update}
                    \parState{Synchronize the codeword assignment matrix $R^{(l)}$ stored outside the training device with the updated $R^{(l)}_B$}
                \EndFor
                
                \State Update learnable parameters $W^{(l)}$ using the estimated gradients $\grad{W^{(l)}}$
            \EndFor
        \EndFor
    \EndProcedure
\end{algorithmic}
\end{algorithm}

\cm{Product VQ and VQ-update rule.}
As described in~\cref{sec:method}, we basically follow the exponential moving average (EMA) update rule for codewords as proposed in~\citep{oord2017neural}. In addition, we propose two further improvements:
\begin{enumerate}[leftmargin=*, noitemsep, topsep=0pt]
    \item Product VQ: we divide the $2f$-dimensional features concatenated with gradients into several $f_\text{prod}$-dimensional blocks. And apply VQ to each of the blocks independently in parallel.
    \item Implicit whitening: we whitening transform the input features and gradients before using them for VQ update, exponentially smooth the mini-batch mean and variance of inputs, and inversely transform the learned codewords using the smoothed mean and variance.
\end{enumerate}

In practice, the product VQ technique is generally required for VQ-GNN to achieve competitive performance across benchmarks and using different GNN backbones. Whitening and Lipschitz regularization are tricks that are helpful in some cases (for example, Lipschitz regularization is only helpful when training GATs). The three techniques are not introduced by us and are already used by some existing work related to VQ but outside of the graph learning community. For example, product VQ is used in~\citep{wu2019learning}, and whitening is used in~\citep{berthelot2018understanding}.

The complete pseudo-code of VQ-update is~\cref{alg:vq-update}. It is important to note that in the experiments, we observed that implicit whitening helps stabilize VQ and makes it more robust across different GNN models and graph datasets. However, we observed that the smoothing of mini-batch mean and variance of gradients (which is used by the approximated backward message passing) is not compatible with some optimization algorithms which consider the cumulative history of gradients, e.g., \emph{Adam}. This practical incompatibility is solved by using \emph{RMSprop} instead of \emph{Adam}

\begin{algorithm}[ht]
\caption{\label{alg:vq-update}VQ-Update: our proposed algorithm to update VQ codewords and assignment using exponential moving average (EMA) estimates of codewords with implicit whitening of inputs.}
\begin{algorithmic}[1]
    \Require Input mini-batch vectors $V\in\sR^{b\times f_\text{prod}}$ as a part of the $2f$-dim feature and gradients
    \Require Exponential decay rate $\gamma$ for momentum estimates of codewords
    \Require Exponential decay rate $\beta$ for momentum estimates of mini-batch mean and variance
    \Require Codewords $\wt{\bar{V}}\in\sR^{k\times f_\text{prod}}$ before update
    \Require Smoothed mean $\wt{\mathrm{E}}[V]$ and variance $\wt{\mathrm{Var}}[V]$ before update
    \Function{VQ-Update$_{(\gamma, \beta)}$}{$V, \wt{\bar{V}}$} \Comment{Update VQ for a $f_\text{prod}$-dim block}
        \State Whitening transform $V$ to $\bar{V}$ and get the mini-batch mean $\mathrm{E}[V]$ and variance $\mathrm{Var}[V]$
        \State $\wt{\mathrm{E}}[V] \gets \wt{\mathrm{E}}[V]\cdot\beta + \mathrm{E}[V]\cdot(1-\beta)$ (EMA update of smoothed mean)
        \State $\wt{\mathrm{Var}}[V] \gets \wt{\mathrm{Var}}[V]\cdot\beta + \mathrm{Var}[V]\cdot(1-\beta)$ (EMA update of smoothed variance)
        \State $R_B \gets $ \Call{FindNearest}{$\bar{V}, \wt{\bar{V}}$} (update assignment, $(R_B)_{i,v}=1$ if $\wt{\bar{V}}_{v,:}$ is closest to $\bar{V}_{i,:}$)
        \State $\ba \gets \ba\cdot\gamma + R_B\tran\vone_b\cdot(1-\gamma)$ (momentum update of cluster sizes)
        \State $\bb \gets \bb\cdot\gamma + R_B\tran \bar{V}\cdot(1-\gamma)$ (momentum update of cluster vector sums)
        \State $\wt{\bar{V}}_{v,:} \gets \frac{1}{\ba_v}\bb_{v,:}$ (update codewords)
        \State Inversely whitening transform $\wt{\bar{V}}$ to $\wt{V}$ using $\wt{\mathrm{E}}[V]$ and $\wt{\mathrm{Var}}[V]$
        \State \Return $\wt{V}$, $R_B$
    \EndFunction
\end{algorithmic}
\end{algorithm}

\cm{Dealing with the row-wise normalized learnable convolutions.}
As mentioned in~\cref{sec:background} and~\cref{apd:generalized-conv}, some self-attention based GNNs including GAT and Graph Transformers require further row-wise normalization of the convolution matrix as $C^{(l,s)}_{i,j} \gets C^{(l,s)}_{i,j}/\sum_j C^{(l,s)}_{i,j}$. However, such a procedure is not characterized by the approximated message passing design in~\cref{sec:method}. Some special treatment is required to normalize the message weights passed to each target node. Actually, this normalization process can be realized by the following three steps:
\begin{enumerate}[leftmargin=*, noitemsep, topsep=0pt]
    \item Padding an extra dimension of ones to the messages, $X^{(l)} \gets X^{(l)}\concat \vone_n$
    \item Message passing using the unnormalized convolution matrix
    \item Normalization by dividing the last dimension, $X^{(l+1)}_{i,1:f} \gets X^{(l+1)}_{i,1:f} / X^{(l+1)}_{i,f+1}$ for any $i=1,\ldots,n$.
\end{enumerate}
In this regard, we decouple the normalization of message weights with the actual message passing process. At the cost of an extra dimension of features (and thus gradients), we can VQ the message passing process again as we did in~\cref{sec:method}. In practice, we found this trick works well, and the experiment results on GAT and Graph Transformer in~\cref{sec:experiments,apd:more-experiments} are obtained using this setup.

\cm{Regularizing the Lipschitz constants of learnable convolutions.}
Since our error bounds on the approximated features and gradients (\cref{thm:feat-error,coro:grad-error}) rely on the Lipschitz constant of learnable convolutions, and the ``decoupled row-wise normalization'' trick discussed above requires some means to control the unnormalized message weights, we have to Lipschitz regularize some learnable convolution GNNs including GAT and Graph Transformers. We follow the approach described in~\citep{dasoulas2021lipschitz} to control the Lipschitz constant of GAT and Graph Transformer without affecting their expressive power. Please see~\citep{dasoulas2021lipschitz} for details.

\cm{Non-linearities, dropout, normalization.}
In experiments, we found our algorithm, VQ-GNN, is compatible with any non-linearities, dropout, and additional batch- or layer-normalization layers.

\cm{Another implementation of VQ-GNN.}
Following recent parallel work, GNNAutoScale~\citep{fey2021gnnautoscale}, it is also possible to implement the VQ-GNN similarly. We can reconstruct the node features of the out-of-mini-batch neighbors for a mini-batch using the learned codewords, and then perform forward-pass message passing between the with-in-mini-batch nodes and the reconstructed nodes. This implementation is more straightforward but may suffer from larger memory overhead when the underlying graph is dense, e.g., on the \textit{Reddit} benchmark.


\section{Implementation Details}
\label{apd:implementation}

\cm{Hardware specs.} Experiments are conducted on hardware with Nvidia GeForce RTX 2080 Ti (11GB GPU), Intel(R) Xeon(R) Silver 4216 CPU @ 2.10GHz, and 32GB of RAM.

\cm{Dataset statistics.}
Detailed statistics of all datasets used in the experiments are summarized in~\cref{tab:dataset-stats}.

\begin{table}[ht]
\centering
\caption{\label{tab:dataset-stats}Information and statistics of the benchmark datasets.}
\begin{tabular}{llllll} \toprule
Dataset        & \textit{ogbn-arxiv}   & \textit{Reddit}       & \textit{PPI}       & \textit{ogbl-collab}  & \textit{Flickr}       \\ \midrule
Task           & node         & node         & node      & link         & node         \\
Setting        & transductive & transductive & inductive & transductive & transductive \\
Label          & single       & single       & multiple  & single       & single       \\
Metric         & accuracy     & accuracy     & F1-score  & hits@50      & accuracy       \\
\# of Nodes    & 169,343      & 232,965      & 56,944    & 235,868      & 89,250       \\
\# of Edges    & 1,166,243    & 11,606,919   & 793,632   & 1,285,465    & 449,878      \\
\# of Features & 128          & 602          & 50        & 128          & 500          \\
\# of Classes  & 40           & 41           & 121       & (2)          & 7            \\
Label Rate     & 54.00\%      & 65.86\%      & 78.86\%   & 92.00\%      & 50.00\%      \\ \bottomrule
\end{tabular}
\end{table}

We note from~\cref{tab:dataset-stats} that: 
\begin{itemize}[leftmargin=*, topsep=1.5pt]
\setlength\itemsep{0.75pt}
    \item \textit{PPI} is a node classification benchmark under the inductive learning setting, i.e., neither attributes nor connections of test nodes are present during training.
    \item \textit{PPI} benchmark comes with multiple labels per node, and the evaluation metric is F1 score instead of accuracy.
    \item \textit{Flickr} and \textit{Reddit} have $500$ and $602$ features per node, respectively. The increased dimensionality of input node features may be challenging for VQ-GNN because VQ has to compress higher-dimensional vectors. Moreover, \textit{Reddit}'s average node degree is $49.8$. More memory is required for mini-batches of the same size because more messages are passed in a layer of GNN.
\end{itemize}

\cm{Hyper-parameter setups of VQ-GNN.} To simplify the settings, we fix the hidden dimension of GNNs to 128 and layer number to 3 across the experiments. We set the size of the VQ codewords to 1024, and its size as small as 256 should also work well. We choose RMSprop (alpha=0.99) as the optimizer, and the learning rate is fixed at 3e-3. To mitigate the error induced by VQ in the high-dimensional space, we split feature vectors into small pieces. In practice, we find that when the dimension of each piece is 4, our algorithm generally works well. When the split dimension is 4 we have 32 separate branches each layer to do the VQ. These branches are independent and can be paralleled. At the end of each layer, separated feature vectors are concatenated together to restore the original hidden dimension, and the restored feature is input to the next layer. Batch norm is used for stable training. We do not use drop out for either our method or the baselines.

\cm{Hyper-parameter setups of other scalable methods.} For baseline models, we follow the practice of OGB. The optimizer is Adam, and the learning rate is fixed at 1e-3. For a fair comparison with respect to memory consumption, on the \texttt{ogbn-arxiv} dataset we set hyper-parameters as below: SAGE-NS with the batch size 85K, per-layer sampling sizes [20,10,5]; ClusterGCN batch size 80, number of partitions 40; GraphSAINT-RW batch size 40K, walk-length 3, number of steps 2. We use the hyper-parameter setting for experiments in Table \ref{tab:efficiency-memory} (the fixed node setting) and Figure \ref{fig:efficiency-converge}. The parameter setting ensures to traverse over all the nodes in the graph in one epoch of training. For fixed message setting in \ref{tab:efficiency-memory}, we alter the batch size of SAGE-NS to 35K, ClusterGCN to 60, GraphSAINT-RW to 60K. Here the batch size for ClusterGCN is small because each batch item means a cluster group of nodes.


\section{Ablation Studies and More Experiments}
\label{apd:more-experiments}

\cm{Experiments on the \textit{Flickr} benchmark}
We conduct another set of performance comparison experiments on the \textit{Flickr} node classification benchmark in~\cref{tab:performance-extra}, whose information and statistics are listed in~\cref{tab:dataset-stats}. As in~\cref{sec:experiments}, we can draw a similar conclusion that VQ-GNN shows more robust performance than the three scalable baselines.
\begin{table}[ht]
\centering
\caption{\label{tab:performance-extra}Performance comparison between sampling-based baselines and our approach, VQ-GNN, on the \textit{Flickr} benchmark.}
\adjustbox{max width=\textwidth}{%
\begin{tabular}{Sl| Sc Sc Sc} \toprule
\multicolumn{1}{c|}{\begin{tabular}[c]{@{}c@{}}Task\\ Benchmark \end{tabular}} & \multicolumn{3}{c}{\begin{tabular}[c]{@{}c@{}}Node Classification (Transductive)\\ \textit{Flickr} (Acc.$\pm$std.) \end{tabular}} \\ \arrayrulecolor{black!50} \midrule
GNN Model               & GCN & SAGE-Mean & GAT \\ \midrule
``Full-Graph''          & $.5209\pm.0053$ & $.5177\pm.0041$ & $.5156\pm.0067$ \\ \arrayrulecolor{black!50} \midrule
NS-SAGE.                & NA\tnote{1} & $.5165\pm.0077$ & $.5282\pm.0052$ \\
Cluster-GCN             & $.4976\pm.0078$ & $.4996\pm.0045$ & $.4907\pm.0107$ \\
GraphSAINT-RW           & $.5239\pm.0071$ & $.5040\pm.0046$ & $.5163\pm.0062$ \\ \arrayrulecolor{black!50} \midrule
\textbf{VQ-GNN (Ours)}  & $.5315\pm.0031$ & $.5323\pm.0083$ & $.5288\pm.0054$  \\ \arrayrulecolor{black} \bottomrule
\end{tabular}}
\end{table}

\cm{Ablation studies}
We also conduct several ablation experiments on the \textit{ogbn-arxiv} benchmark with GCN backbone. Please note that if not mentioned otherwise, the hyper-parameter setups will follow the corresponding model and are described in~\cref{apd:implementation}.

\begin{itemize}[leftmargin=*, topsep=1.5pt]
\setlength\itemsep{0.75pt}
    \item \textbf{Performance vs. the number of layers}: We vary the number of layers of GCN backbone from one to five, and the performance of VQ-GNN on \textit{ogbn-arxiv} is shown in the following table. We see that a two-layer GCN is sufficient to achieve good performance on \textit{ogbn-arxiv} while using only one layer harms the performance. Stacking more layers will not bring any performance gain.
\begin{table}[H]
\begin{tabular}{llllll} \toprule
\# of Layers & 1      & 2      & 3      & 4      & 5      \\ \midrule
Accuracy     & 0.6599 & 0.7006 & 0.7055 & 0.7080 & 0.7037 \\ \bottomrule
\end{tabular}
\end{table}
    
    \item  \textbf{Performance vs. codebook size}: We vary the codebook size from 64 to 4096 and fix the mini-batch size at 40K. Performance is shown as follows, where we can see the performance is not sensitive to the codebook size. Setting the codebook size to 64 is enough to achieve relatively good performance on \textit{ogbn-arxiv}. This may indicate that the node feature distribution of \textit{ogbn-arxiv} is sparse, which is reasonable as the 128-dimensional node features are obtained by averaging the embedding of words in the titles and abstracts of arXiv papers~\citep{hu2020open}.
\begin{table}[H]
\begin{tabular}{lllll} \toprule
Codebook Size & 64     & 256    & 1024   & 4096   \\ \midrule
Accuracy      & 0.6950 & 0.7011 & 0.7030 & 0.7049 \\ \bottomrule
\end{tabular}
\end{table}

    \item \textbf{Performance vs. mini-batch size}: We vary the mini-batch size from $10$K to $80$K, and the performance is shown as follows. We see that smaller mini-batch sizes can slightly lower the overall performance. When the mini-batch size is smaller, more messages come from out-of-mini-batch nodes (see~\cref{fig:schema-graph}), and they are approximated by the VQ codewords. This increased number of approximated messages can harm the performance. However, generally speaking, the performance is not sensitive to the mini-batch size as long as it is not small.
\begin{table}[H]
\begin{tabular}{llllll} \toprule
Mini-batch Size & 10K    & 20K    & 40K    & 60K    & 80K    \\ \midrule
Accuracy        & 0.6843 & 0.6920 & 0.7011 & 0.7055 & 0.7061 \\ \bottomrule
\end{tabular}
\end{table}

    \item \textbf{Performance vs. mini-batch sampling strategy}: We compare three different sampling strategies: (1) randomly sampling nodes, (2) randomly sampling edges, and (3) sampling using random walks as in GraphSAINT-RW. As shown in the following table, we did not observe an obvious difference in the performance.
\begin{table}[H]
\adjustbox{max width=\textwidth}{%
\begin{tabular}{llll} \toprule
Mini-batch Sampling Techniques & Sampling Nodes & Sampling Edges & Sampling using Random Walks \\ \midrule
Accuracy                       & 0.7020         & 0.7034         & 0.7023                      \\ \bottomrule
\end{tabular}}
\end{table}
\end{itemize}

\cm{Performance of VQ-GNN with Graph Transformers~\citep{yaronlipman2020global}.} Our algorithm enables graph transformer architectures to compute global attention on large graphs. For each layer, we input hidden features into VQ-GNN, Graph Transformer, and Linear modules separately and sum the output features of each module together. In this way, the holistic model can not only leverage global attention but can also absorb local information. The transformer module is adapted from \citep{yaronlipman2020global}. We show the performance in Table \ref{tab:performance-powerful}. We find that the removal of the batch norm will mitigate the problem of overfitting, so our model does not involve batch norm.

\begin{table}[ht]
\centering
\caption{\label{tab:performance-powerful}Performance of VQ-GNN with Graph Transformer Backbone on the \textit{ogbn-arxiv} benchmark.}
\begin{tabular}{Sl | Sc Sc Sc Sc} \toprule
\multicolumn{1}{l|}{Benchmark} & \multicolumn{1}{l}{\textit{ogbn-arxiv} (Acc.$\pm$std.)}  \\ \midrule
Global Attention $+$ GAT~\citep{yaronlipman2020global} & $.7108\pm.0055$  \\ \bottomrule
\end{tabular}
\end{table}







\end{document}